\documentclass{article}

\usepackage{arxiv}
\usepackage[utf8]{inputenc}
\usepackage[T1]{fontenc}
\usepackage{hyperref}
\usepackage{url}
\usepackage{booktabs}
\usepackage{amsfonts}
\usepackage{amsmath}
\usepackage{amssymb}
\usepackage{graphicx}
\usepackage{microtype}
\usepackage{array}
\usepackage{multirow}
\usepackage{enumitem}

\title{From Surface Forecasting to Observability Forecasting:\\ A Latent World Model for Cloud-Aware EO Monitoring}

\author{
  Mohanad Albughdadi \\
  European Centre for Medium-Range Weather Forecasts \\
  \texttt{mohanad.albughdadi@ecmwf.int}
}

\begin{document}
\maketitle

\begin{abstract}
The bottleneck of Earth Observation processing chains is not the arrival of new imagery but whether the surface is actually visible when the image arrives. We study this as an \emph{observability forecasting} problem on \texttt{EarthNet2021}. Given recent multispectral imagery and exogenous weather drivers, the goal is to predict whether the next acquisition will be usable and, if not, when a usable view is likely to return. To do this, we adapt LeWorldModel, a joint-embedding predictive architecture world model, to cloud-aware Earth Observation sequences. The final pipeline converts raw minicubes into episodic HDF5 sequences with five image channels (blue, green, red, near-infrared, cloud mask) and eight meteorological and calendar covariates. The resulting model has $18.0$M trainable parameters and is trained from scratch on $23,904$ training episodes.

The trained leWorldModel is evaluated under a locked protocol: linear probes are fit on train only, calibration choices are set on an internal validation split, and the fitted heads are then frozen for \texttt{valsplit}, IID, OOD, and extreme evaluation. On the full frozen-bundle observability benchmark, LeWorldModel consistently outperforms persistence. For next-step usability, balanced accuracy ranges from $0.769$ to $0.887$, compared with $0.493$ to $0.556$ for persistence. For exact first-usable-horizon prediction, accuracy ranges from $0.602$ to $0.806$, compared with $0.120$ to $0.369$ for persistence. Against a frozen LightGBM baseline fit on the same training windows, LeWorldModel is better on continuous clear/cloud regression and on exact recovery timing on \texttt{valsplit}, IID, and extreme, while LightGBM is stronger on the simpler binary any-usable-within-six task and is more robust on OOD. In separate sampled diagnostic analyses, LeWM also produces strong ranking-based anomaly signals under synthetic temporal inconsistencies. Overall, the model is most convincing here as a latent observability forecaster for cloud-aware Earth Observation monitoring, not as a spatial video predictor or a planning model. Model's code, weights, and experiments are available on this repository \url{https://github.com/AlbughdadiM/lewm-eo-cloud-monitoring}.
\end{abstract}


\section{Introduction}
Earth Observation~(EO) processing pipelines do not fail because imagery stops arriving. They fail because the images that arrive cannot be used. Clouds, haze, and missing valid pixels can hide the surface for days or weeks, making the observation quality the main challenge rather than data availability. In operational settings, that changes the question. The issue is not only what the surface will look like next, but whether the next acquisition will be usable at all and, if not, how long the wait will be.

We treat that as an \emph{observability forecasting} problem. The aim is not pixel reconstruction, surface-index regression, or planning. It is to estimate when the scene becomes observable again after cloud-obstructed acquisitions.

\texttt{EarthNet2021} was introduced as a benchmark for forecasting high-resolution Earth surface observations from coarser meteorological drivers \cite{requena2021earthnet2021}. Most work on the dataset therefore focuses on video prediction or surface forecasting. We use the same dataset for a different question: can a latent world model forecast whether future observations will be usable, whether any usable observation will return within a short horizon, and when that return is most likely to happen?

LeWorldModel~(LeWM) is a recent world-model architecture based on a Joint-Embedding Predictive Architecture~(JEPA) \cite{maes2026leworldmodel}. Instead of reconstructing pixels, it learns a latent state, predicts future latent states from past latents and covariates, and regularizes that space toward a stable Gaussian geometry. LeWM is a sensible model for cloud-aware EO monitoring because observability is fundamentally a temporal state problem, not a pixel synthesis problem. What matters operationally is not whether the model can draw the next image, but whether it can track the hidden evolution of scene visibility under changing weather conditions and recent observation history. \texttt{EarthNet2021} provides exactly the ingredients for that setting: irregularly useful observations, cloud-driven partial observability, and exogenous meteorological covariates that shape what can be seen next. A JEPA world model is well matched to this structure. It learns a predictive state that must carry forward the information needed to anticipate future observation quality, while avoiding the burden of reconstructing every spatial detail of a cloudy frame. That is an important distinction. For this task, exact pixels are often ambiguous or operationally irrelevant, whereas the latent transition from obscured to usable is the signal that matters. The appeal of LeWM, then, is not just that it is a modern latent predictor; it is that its training objective is closer to the decision problem posed by cloud-aware monitoring.

The research question addressed in this paper is straightforward: once LeWM is adapted carefully to \texttt{EarthNet2021} and evaluated under a locked protocol, does it provide a useful signal for cloud-aware monitoring?

Our contributions are practical and empirical:
\begin{itemize}[leftmargin=1.4em]
\item We reformulate the \texttt{EarthNet2021} use case from generic surface forecasting to \emph{observability forecasting under cloud occlusion}, with explicit targets for next-step usability, any usable future observation within horizon, and first usable future horizon.
\item We provide a reproducible LeWM adaptation pipeline for this setting, including cloud-aware conversion from raw minicubes to episodic HDF5, explicit cloud-mask handling, and dataset-specific normalization.
\item We define a locked evaluation protocol that separates model training, linear-probe fitting, validation-time calibration, and final reporting across \texttt{EarthNet2021} \texttt{valsplit}, IID, OOD, and extreme splits.
\item We show that LeWM is most convincing not as a pixel forecaster but as an observability model: it improves over persistence in latent forecasting, strongly predicts whether future observations will be usable, estimates when usable observations return, and provides useful temporal anomaly-ranking signals.
\item We compare LeWM against both persistence and a frozen LightGBM baseline. The comparison reveals a clear task split: LeWM is strongest on continuous observation-quality regression and on exact recovery timing on most splits, while LightGBM is strongest on the simpler binary any-usable-within-horizon task and on OOD more broadly.
\end{itemize}

In short, the value of LeWM on \texttt{EarthNet2021} is not that it becomes a spatial forecaster or a planning system. The value is that it can act as a latent observability forecaster for cloud-aware EO monitoring.

\section{Related Work}
\paragraph{Earth surface forecasting with EarthNet2021 and related EO forecasters.}
\texttt{EarthNet2021} was introduced as a benchmark for forecasting high-resolution Earth surface observations conditioned on mesoscale weather drivers \cite{requena2021earthnet2021,requenamesa2020earthnet}. In that literature, the main objective is future surface prediction, either as image forecasting or as prediction of downstream surface indicators from recent satellite context and future forcing. Earthformer broadened that agenda by showing that transformer-based spatiotemporal models can capture geophysical dynamics at scale \cite{gao2022earthformer}. EO-WM pushes further toward probabilistic EO forecasting with weather-aware conditioning and diagnostics for weather-response fidelity \cite{luo2026eowm}. We keep the same \texttt{EarthNet2021} setting, but change the question. Our target is not future surface reconstruction. It is whether the future observation itself will be usable and when a usable view will return.

\paragraph{Clouds, missing data, and cloud-gap reconstruction.}
Clouds and invalid pixels are a routine obstacle in EO monitoring, so much of the related literature treats the problem as gap filling, imputation, or cloud removal. SEN12MS-CR-TS formalized multimodal, multitemporal cloud removal with optical and SAR context \cite{ebel2022sen12mscrts}. UnCRtainTS adds calibrated uncertainty to cloud removal in optical satellite time series \cite{ebel2023uncrtaints}. Other work uses partial convolutions or foundation-model-based priors for cloud-gap imputation \cite{appel2024efficient,godwin2024seeing}. These methods are directly relevant as context, but they solve a different task. Their goal is to reconstruct the hidden surface. Ours is to forecast observation usability itself: whether the next acquisitions are useful enough for monitoring, and how long recovery will take.

\paragraph{EO self-supervised learning and foundation models.}
A parallel line of work focuses on EO-specific representation learning rather than explicit future forecasting. SatMAE adapts masked autoencoding to temporal and multispectral satellite imagery \cite{cong2022satmae}, while Scale-MAE emphasizes multiscale geospatial representation learning \cite{reed2022scalemae}. Larger geospatial foundation-model efforts such as Prithvi argue that EO data benefit from domain-specific pretraining and broad transfer across downstream tasks \cite{jakubik2023generalistgeoai}. Recent work further suggests that EO representation learning benefits from explicit geo-temporal conditioning and does not necessarily require very large models: a lightweight metadata-aware Mixture-of-Experts masked autoencoder combines imagery with latitude/longitude and cyclic seasonal/daily encodings and achieves competitive frozen-encoder transfer with only 2.5M parameters \cite{albughdadi2025lightweight}. Together, these studies support the broader premise that EO representation learning should not simply inherit assumptions from natural-image pretraining.

\paragraph{Foundation models versus latent world models.}
Even so, EO foundation models and latent world models are not interchangeable. Most EO foundation-model papers evaluate transfer to static or weakly temporal downstream tasks such as classification, segmentation, or regression after fine-tuning or linear probing. A world model asks for something more structured: a state representation that evolves coherently under time and exogenous forcing. That difference is central in our setting. Forecasting cloud-aware observability is not just a question of whether a representation is useful downstream. It is a question of whether the representation preserves the temporal state needed to predict visibility, usability, and recovery timing.

\paragraph{Latent predictive models, JEPA, and world models.}
Outside EO, a long line of work studies predictive latent dynamics from high-dimensional observations. World Models, PlaNet, and Dreamer showed that compact latent states can support multi-step prediction and planning without reconstructing every pixel at every step \cite{ha2018worldmodels,hafner2018planet,hafner2019dreamer}. JEPA-style learning shifts the emphasis from pixel reconstruction to representation prediction \cite{assran2023ijepa}. V-JEPA 2 extends that view to video understanding, prediction, and planning \cite{assran2025vjepa2}. LeWorldModel brings these ideas together in an end-to-end JEPA world model with latent prediction and Gaussian regularization \cite{maes2026leworldmodel}. That makes it a natural candidate for EO settings where distinct pixel realizations can still correspond to the same monitoring outcome. In our case, different future images can all lead to the same answer: the observation remains unusable because cloud cover stays high.

\paragraph{Position of this work.}
This paper sits at the intersection of these lines of work. From \texttt{EarthNet2021} and EO forecasting, it inherits multispectral image sequences driven by exogenous weather. From cloud-gap work, it inherits the practical fact that cloud cover often sets the real limit on EO monitoring. From EO representation learning, it inherits the view that EO data benefit from domain-specific pretraining. From JEPA and world-model work, it inherits the idea that predictive state can be learned in latent space without explicit future-image synthesis. What is still missing in prior work is an explicit treatment of observability forecasting: whether the next observation will be usable, whether any usable observation is likely to appear within a short horizon, and when the first usable future observation is likely to return. That is the gap addressed here.
\section{Application Setting}
We frame the target use case as follows. An EO monitoring pipeline receives a sequence of multispectral observations and associated exogenous weather drivers. At each time step, the pipeline may need to decide among actions such as:
\begin{itemize}[leftmargin=1.4em]
\item whether the next acquisition is likely to be usable enough for surface analysis;
\item whether to defer a monitoring decision because the next few observations will likely remain cloud-obstructed;
\item when a useful observation is likely to become available again;
\item whether a newly observed transition is temporally inconsistent with the recent context and should be flagged for quality control.
\end{itemize}

These are forecasting and inference problems, not control problems. The weather variables in \texttt{EarthNet2021} are \emph{exogenous covariates}, not actions that a planner can optimize. This distinction matters. It is valid to test forecast sensitivity to perturbed weather inputs, but it is not valid to claim policy optimization or action planning from this dataset alone.

\section{Data and Preparation}
\subsection{Raw EarthNet2021 structure}
\texttt{EarthNet2021} training minicubes contain high-resolution Sentinel-2 image sequences and mesoscale meteorological variables \cite{requena2021earthnet2021}. In the raw training split, each cube contains:
\begin{itemize}[leftmargin=1.4em]
\item \texttt{highresdynamic} of shape $(128,128,7,30)$;
\item \texttt{mesodynamic} of shape $(80,80,5,150)$;
\item static variables that are not used in the present LeWM adaptation.
\end{itemize}

For the high-resolution dynamic tensor, the relevant channels are:
\begin{itemize}[leftmargin=1.4em]
\item channels $0\ldots3$: blue, green, red, near-infrared (NIR);
\item channel 6: binary cloud mask (\texttt{cldmask}), interpreted as 1 = cloudy and 0 = clear.
\end{itemize}
Channels related to cloud probability and scene class are not used by the final model. For the external paired context/target splits, the stored schema is already simplified: each dynamic cube has shape $(128,128,5,T)$ with the binary cloud mask in channel 4.

\subsection{Cloud-aware conversion to LeWM episodes}
We convert each minicube into one episodic sequence in the LeWM HDF5 format. The final observation tensor keeps five channels,
\[
x_t \in \mathbb{R}^{128\times128\times5} = [\text{blue},\text{green},\text{red},\text{nir},\text{cloud\_mask}],
\]
while the covariate vector is
\[
u_t = [\Delta t,\sin(\mathrm{doy}),\cos(\mathrm{doy}),\mathrm{rain},\mathrm{pressure},\mathrm{temp\_mean},\mathrm{temp\_min},\mathrm{temp\_max}] \in \mathbb{R}^{8}.
\]

A central design decision is to \emph{preserve all time steps} while making cloud occlusion explicit. For each frame we:
\begin{enumerate}[leftmargin=1.6em]
\item clip the four spectral bands to $[0,1]$ following the \texttt{EarthNet2021} convention;
\item set cloudy or non-finite spectral pixels to zero;
\item append the binary cloud mask as an additional channel instead of dropping cloudy frames.
\end{enumerate}
This choice preserves the temporal alignment between satellite imagery and weather covariates. It also lets the model distinguish \emph{true low reflectance} from \emph{missing surface information due to clouds}. In the final training configuration we keep all frames, so each converted training episode remains 30 steps long.

\subsection{Time features and weather aggregation}
\texttt{EarthNet2021} weather is stored at higher temporal resolution than the image sequence. For every retained transition from frame $t$ to frame $t+1$, we compute:
\begin{itemize}[leftmargin=1.4em]
\item the day gap $\Delta t$ between the two image timestamps;
\item cyclical day-of-year features $\sin(2\pi\,\mathrm{doy}/365.25)$ and $\cos(2\pi\,\mathrm{doy}/365.25)$;
\item interval weather statistics from \texttt{mesodynamic}.
\end{itemize}
Precipitation is aggregated by summation over the transition interval. Pressure and temperature variables are aggregated by averaging over the same interval. In effect, the action/covariate vector describes the exogenous forcing between successive image observations.

\subsection{Training HDF5 and normalization}
The final training HDF5 contains 717,120 steps and 23,904 episodes. Every episode has length 30 and every observation uses the native $128\times128$ spatial resolution. The evaluation splits are converted with the same representation, with the extreme split yielding 60-step episodes (see Table~\ref{tab:data}).

\begin{table}[!htp]
\small
\centering
\caption{Dataset summary after conversion to episodic HDF5, derived directly from the converted split files. \texttt{valsplit} is an internal train/validation window split, not a separate HDF5 file.}
\label{tab:data}
\begin{tabular}{lrrrrrr}
\toprule
Split & Steps & Episodes & Median ep. len & Valid frac & Clear frac & Cloud frac \\
\midrule
Train source & 717120 & 23904 & 30 & 0.982 & 0.596 & 0.404 \\
IID & 126570 & 4219 & 30 & 0.982 & 0.593 & 0.407 \\
OOD & 126420 & 4214 & 30 & 0.979 & 0.609 & 0.391 \\
Extreme & 240000 & 4000 & 60 & 0.968 & 0.313 & 0.687 \\
\bottomrule
\end{tabular}
\end{table}

We estimate per-channel pixel normalization statistics directly from the converted \texttt{EarthNet2021} training data, using only clear spectral pixels that are not all-zero after masking. This avoids importing image statistics from unrelated natural-image datasets. The cloud mask channel is left binary with mean 0 and standard deviation 1 (see Table~\ref{tab:stats}).

\begin{table}[!htp]
\small
\centering
\caption{Per-channel normalization used in the final training run, derived from the converted training HDF5 statistics. Spectral statistics are estimated from the converted train split; the cloud mask remains binary.}
\label{tab:stats}
\begin{tabular}{lrr}
\toprule
Channel & Mean & Std \\
\midrule
blue & 0.0522 & 0.0323 \\
green & 0.0797 & 0.0416 \\
red & 0.0882 & 0.0643 \\
nir & 0.2899 & 0.0896 \\
cloud\_mask & 0.0000 & 1.0000 \\
\bottomrule
\end{tabular}
\end{table}

\section{Model}
\subsection{EarthNet2021 adaptation of LeWorldModel}
We keep the original LeWM training path and adapt only the input interface and dataset pipeline \cite{maes2026leworldmodel}. The instantiated model has 18.0M trainable parameters and consists of:
\begin{itemize}[leftmargin=1.4em]
\item a $1\times1$ learned channel adapter from the 5 \texttt{EarthNet2021} channels to the 3-channel ViT encoder input expected by the backbone;
\item a ViT-tiny encoder with patch size 16 and image size 128;
\item an MLP projector that maps the encoder class token to a 192-dimensional latent state;
\item an MLP action encoder that maps the 8-dimensional covariate vector to the same latent width;
\item a 6-layer conditional autoregressive transformer predictor with 16 heads.
\end{itemize}

Only the encoder class token is used for the forecasting path of the final model. Patch tokens are exposed in code for probing, but they are not part of the forecasting loss used in the reported checkpoint.

\subsection{State and covariate notation}
Let $x_t$ be the multispectral-cloud observation and let $u_t$ be the covariate vector. The encoder and action encoder produce
\[
z_t = E_\theta(x_t) \in \mathbb{R}^{192}, \qquad a_t = A_\phi(u_t) \in \mathbb{R}^{192}.
\]
For a training window of length 11 with history size 10, the model observes $(z_1,\ldots,z_{10})$ and $(a_1,\ldots,a_{10})$ and predicts the next latent state at every position:
\[
(\hat z_2,\ldots,\hat z_{11}) = P_\psi(z_1,\ldots,z_{10}, a_1,\ldots,a_{10}).
\]
This is important: the final training loss is \emph{teacher-forced over the whole context window}, not only at the terminal step.

\subsection{Training objective}
The final objective is
\begin{equation}
\mathcal{L} = \mathcal{L}_{\mathrm{pred}} + \lambda_{\mathrm{sig}}\,\mathcal{R}_{\mathrm{sig}}(z_{1:11}),
\end{equation}
with
\begin{equation}
\mathcal{L}_{\mathrm{pred}} = \frac{1}{10D}\sum_{t=1}^{10}\lVert \hat z_{t+1}-z_{t+1}\rVert_2^2.
\end{equation}
$\mathcal{R}_{\mathrm{sig}}$ is the LeWM isotropic-Gaussian regularizer based on random projections.
\section{Training Setup}
The reported checkpoint is trained from scratch with AdamW on a single NVIDIA A100 20GB MIG instance. The final run uses batch size 32, learning rate $3\times10^{-4}$, weight decay $10^{-3}$, bf16 mixed precision, and 24 epochs. Gradient clipping is set to 1.0 and $\lambda_{\mathrm{sig}}=0.05$. The used training configuration are available in Table~\ref{tab:train}.
\begin{table}[!htp]
\small
\centering
\caption{Final training configuration for the reported LeWM checkpoint, taken from the frozen run configuration.}
\label{tab:train}
\begin{tabular}{ll}
\toprule
Item & Value \\
\midrule
History size & 10 \\
Number of future predictions & 1-step targets over 10 teacher-forced positions \\
Latent width & 192 \\
Encoder & ViT-tiny, patch size 16 \\
Batch size & 32 \\
Optimizer & AdamW \\
Learning rate & $3\times 10^{-4}$ \\
Weight decay & $10^{-3}$ \\
Precision & bf16-mixed \\
Gradient clip & 1.0 \\
Epochs & 24 \\
SIGReg weight & 0.05 \\
\bottomrule
\end{tabular}
\end{table}

\begin{figure}[!htp]
  \centering
  \includegraphics[width=\linewidth]{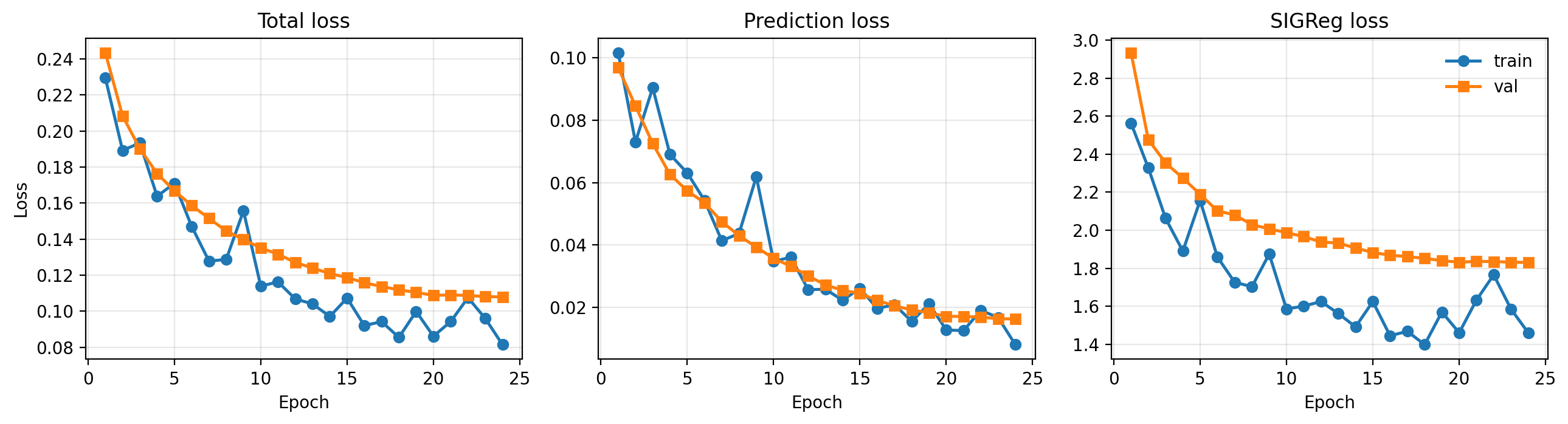}
  \caption{LeWM training and validation curves on \texttt{EarthNet2021}, plotted from the final run log used for downstream evaluation.}
  \label{fig:training}
\end{figure}

Figure~\ref{fig:training}, plotted from the final training log, shows that training is stable and converges smoothly. Both the total loss and the latent prediction loss decrease throughout training, while the validation regularization term remains controlled. The final validation prediction loss is 0.0164 and the final validation SIGReg term is 1.83.

\section{Evaluation Protocol}
\subsection{Window definitions}
With history size 10 and one future target step, every 30-step episode yields 20 standard next-step evaluation windows. For six-step rollout evaluation, every 30-step episode yields 15 windows, because six additional future steps must be available after the context. The extreme split contains 60-step episodes and therefore yields 50 standard windows and 45 rollout windows per episode.

The full training HDF5 is internally split 90/10 into train and validation windows. This produces 430,273 train windows and 47,807 validation windows for the standard next-step task. The script-based benchmark traverses all available windows on the external evaluation splits, while the notebook diagnostics use smaller fixed subsamples for tractability.

\subsection{Locked downstream protocol}
We use one protocol across all reported splits:
\begin{enumerate}[leftmargin=1.6em]
\item train LeWM once on the \texttt{EarthNet2021} training HDF5;
\item fit all linear probes on training windows only;
\item tune minor calibration choices on \texttt{valsplit} only;
\item freeze those probes and thresholds for IID, OOD, and extreme.
\end{enumerate}
This matters because refitting a downstream head separately on each test split would blur the distinction between true generalization and split-specific adaptation.

In the results that follow, the observability benchmark uses the frozen full-bundle outputs, whereas latent forecasting, latent geometry, anomaly ranking, and weather sensitivity come from those fixed notebook diagnostic windows.

The paper reports two evaluation modes: \textit{i)} the full evaluation frozen-bundle on all available \texttt{valsplit}, IID, OOD, and extreme windows to benchmark observability experiments, as well as \textit{ii)} smaller fixed budgets for latent forecasting, latent geometry, anomaly ranking, and weather sensitivity. This separation keeps the reported benchmark exhaustive while keeping the exploratory figures tractable.

\subsection{Baselines}
Two baselines are used to compare their results with the results of LeWM:
\begin{itemize}[leftmargin=1.4em]
\item \textbf{Persistence.} For latent forecasting, predict every future latent by copying the last observed latent state. For the observation-quality tasks, persistence instead uses the last observed quality score, i.e. the minimum of clear fraction and valid fraction in the final context frame.
\item \textbf{LightGBM observability baseline.} A frozen tabular baseline fit once on the train-side windows of the converted \texttt{EarthNet2021} HDF5. It uses only explicit quality histories and future weather/calendar covariates, not learned image latents. This baseline is deliberately strong for binary observability decisions and therefore helps separate what truly requires a latent world model from what can be solved well with structured tabular signals.
\end{itemize}

\subsection{Tasks and their monitoring objective}
We evaluate five task families. The first two are supporting diagnostics: they verify that the latent dynamics model has learned non-trivial temporal structure beyond persistence. The last three are the application-facing tasks that motivate the paper: anomaly ranking for temporal quality assurance, next-step observation usability, and forecasting when usable observations return.

\paragraph{1. Next-step latent forecasting.}
\emph{Objective:} determine whether the model learned non-trivial one-step dynamics at all. We compare the predicted next latent against the true next latent and against persistence.

\paragraph{2. Six-step autoregressive rollout.}
\emph{Objective:} test whether the latent dynamics remain useful when the model is rolled forward recursively. Future weather covariates from the dataset are provided during rollout, so this is forecasting under known exogenous forcing rather than unconditional image generation.

\paragraph{3. Ranking-based anomaly detection.}
\emph{Objective:} flag temporally inconsistent futures for quality control. We synthesize three anomaly types: future-frame shuffle, future-action shuffle, and future visual shift. The anomaly score is the one-step latent prediction error. We report AUROC, average precision, and score ratio.

\paragraph{4. Next-step observation quality.}
\emph{Objective:} estimate whether the next acquisition will be usable enough for surface analysis. We fit linear probes from the predicted next latent to next-step clear fraction, next-step cloud fraction, and a binary \texttt{usable\_next} target. A step is usable when both clear fraction and valid fraction are at least 0.8.

\paragraph{5. First usable future observation.}
\emph{Objective:} estimate when monitoring can resume after cloudy or low-validity observations. Using six-step rollout latents, we predict both whether any usable observation occurs within the horizon and the exact horizon of the first usable step.

We additionally run a \emph{weather-sensitivity analysis} by perturbing the last observed weather covariates ($\pm 5$ mm rain, $\pm 2^{\circ}$C temperature). Because \texttt{EarthNet2021} actions are exogenous, we interpret this as sensitivity analysis, not planning.

\section{Results}
\subsection{Latent forecasting as supporting evidence}
\begin{figure}[!htp]
  \centering
  \includegraphics[width=\linewidth]{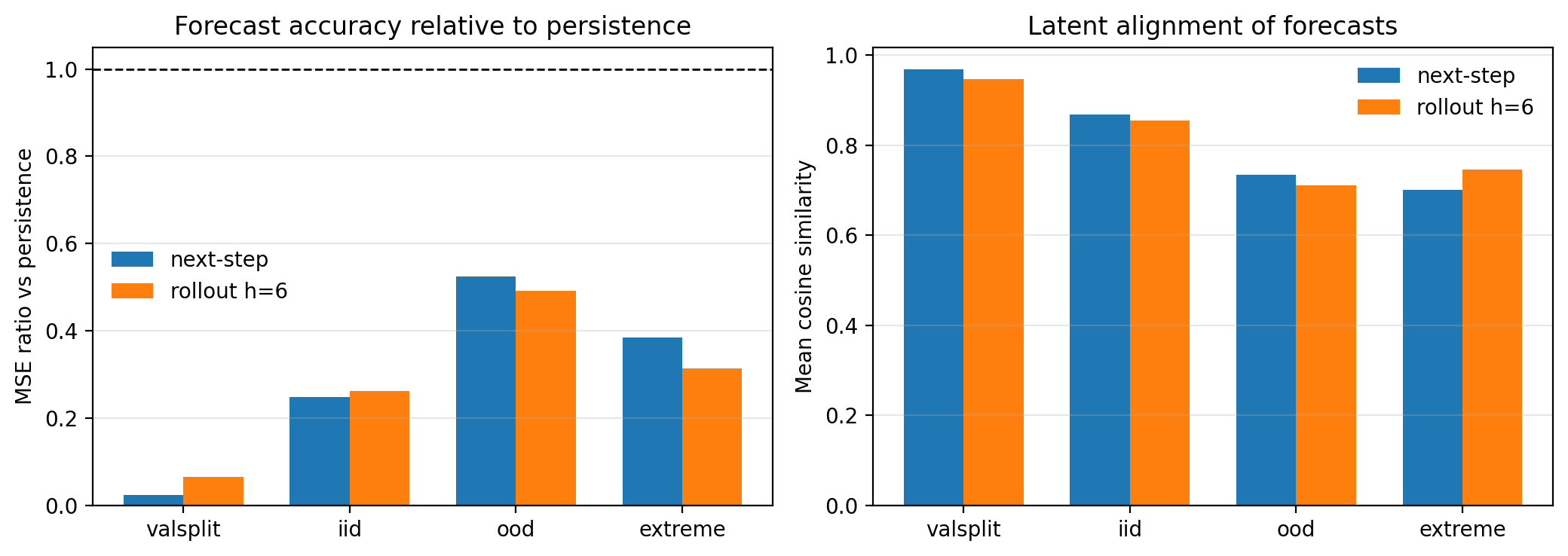}
  \caption{Sampled latent-forecast diagnostics across splits from the executed notebooks. Left: mean-squared-error ratio relative to persistence (lower is better; dashed line is parity with persistence). Right: cosine similarity between predicted and target latent states. LeWM remains better than persistence on every split, but the margin narrows substantially on OOD.}
  \label{fig:forecast}
\end{figure}

Before interpreting the observability tasks, we verified on fixed sampled diagnostic windows that the latent world model actually learns predictive temporal structure rather than simply copying the last state, and this was the case on every split. Figure~\ref{fig:forecast} shows that LeWM improves over persistence for both next-step prediction and six-step rollout. The effect is strongest on \texttt{valsplit}, degrades on IID, weakens further on OOD, and partially recovers on extreme.

This ordering is informative. The extreme split is harder in cloud frequency and sequence length, but its latent forecasts remain better than OOD on several metrics. The main limitation is therefore not simply longer horizons or more clouds; it is distribution shift in the latent dynamics.

\begin{figure}[!htp]
  \centering
  \includegraphics[width=\linewidth]{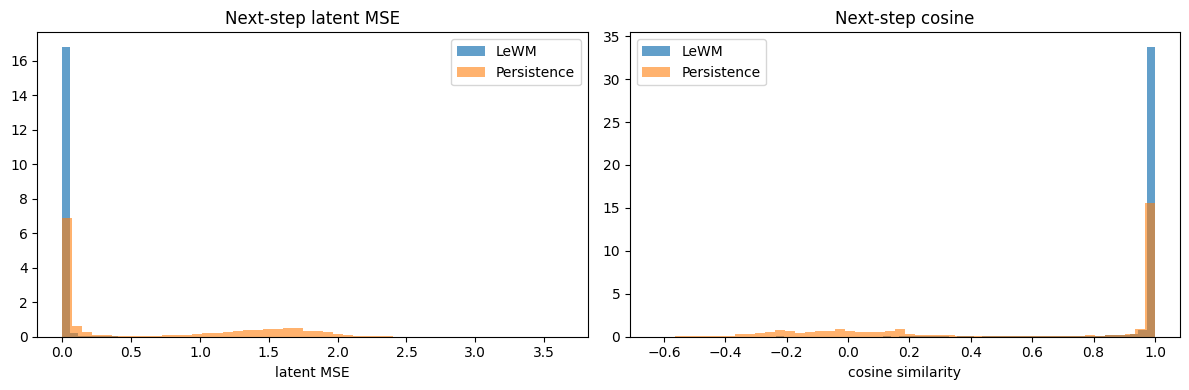}
  \caption{Detailed next-step diagnostics from a fixed sampled subset of the \texttt{valsplit}. Left: histogram of latent MSE. Right: histogram of latent cosine similarity. These distributions explain the strong average next-step metrics: LeWM places much more mass in the near-perfect prediction regime than persistence.}
  \label{fig:nextstepdetail}
\end{figure}

\begin{figure}[!htp]
  \centering
  \includegraphics[width=\linewidth]{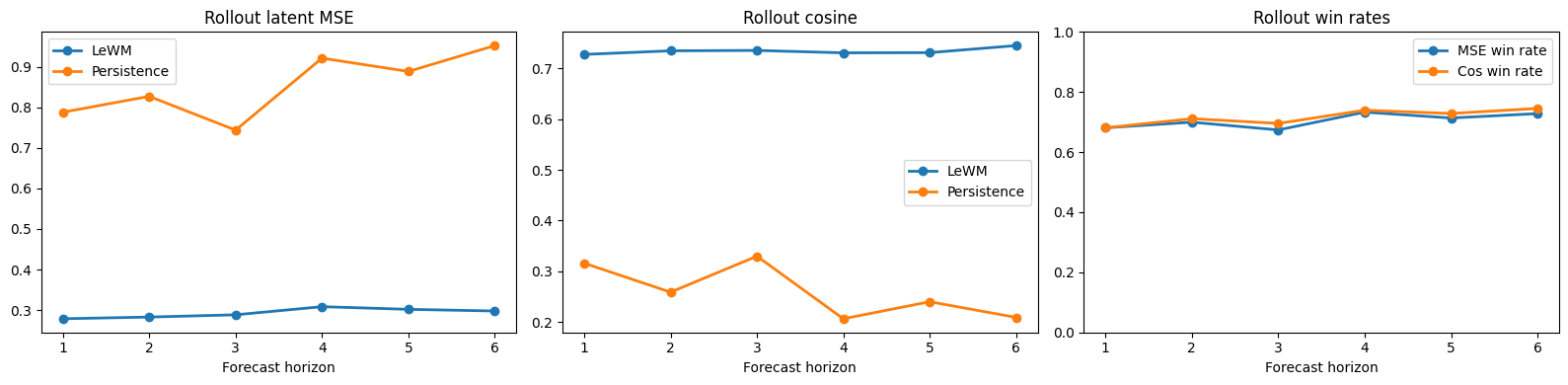}
  \caption{Detailed rollout diagnostics from a fixed sampled subset of the extreme split. LeWM maintains lower rollout MSE, higher cosine similarity, and better per-window win rates than persistence throughout the full six-step forecast horizon.}
  \label{fig:extremerollout}
\end{figure}

To make the aggregate improvements more concrete, Figure~\ref{fig:nextstepdetail} shows the full held-out next-step error distributions on \texttt{valsplit}. The LeWM predictions are concentrated near zero latent error and cosine similarity one, whereas persistence produces a broad high-error tail. Figure~\ref{fig:extremerollout} shows the detailed horizon-by-horizon rollout curves on the extreme split. Even in this much cloudier and longer-horizon regime, LeWM remains better than persistence at every horizon.

\subsection{Latent geometry}
The Gaussian regularizer forces an isotropic regime on the latent dynamics, especially on train-like data and degrades gracefully under shift. On the fixed sampled notebook diagnostics summarized in Table~\ref{tab:geometry}, \texttt{valsplit} and IID remain near the intended isotropic regime, with median per-dimension standard deviation close to one and modest off-diagonal covariance. OOD shifts the latent mean away from zero. Extreme compresses the latent variance considerably, which is consistent with its much lower clear-fraction statistics.

\begin{table}[!htp]
\small
\centering
\caption{Latent geometry diagnostics from fixed sampled held-out encoded vectors in the executed split notebooks. Lower mean absolute mean and covariance, and median standard deviation near one, indicate a healthier approximately isotropic latent space.}
\label{tab:geometry}
\begin{tabular}{lrrrr}
\toprule
Split & $\mathrm{mean}|\mu_d|$ & median $\sigma_d$ & mean $\mathrm{var}_d$ & mean $|\mathrm{cov}_{ij}|$ \\
\midrule
Valsplit & 0.0471 & 1.0445 & 1.1176 & 0.2112 \\
IID & 0.0445 & 1.0340 & 1.1012 & 0.2070 \\
OOD & 0.1244 & 1.0228 & 1.1133 & 0.2491 \\
Extreme & 0.1808 & 0.6965 & 0.5527 & 0.1709 \\
\bottomrule
\end{tabular}
\end{table}

\begin{figure}[!htp]
  \centering
  \includegraphics[width=\linewidth]{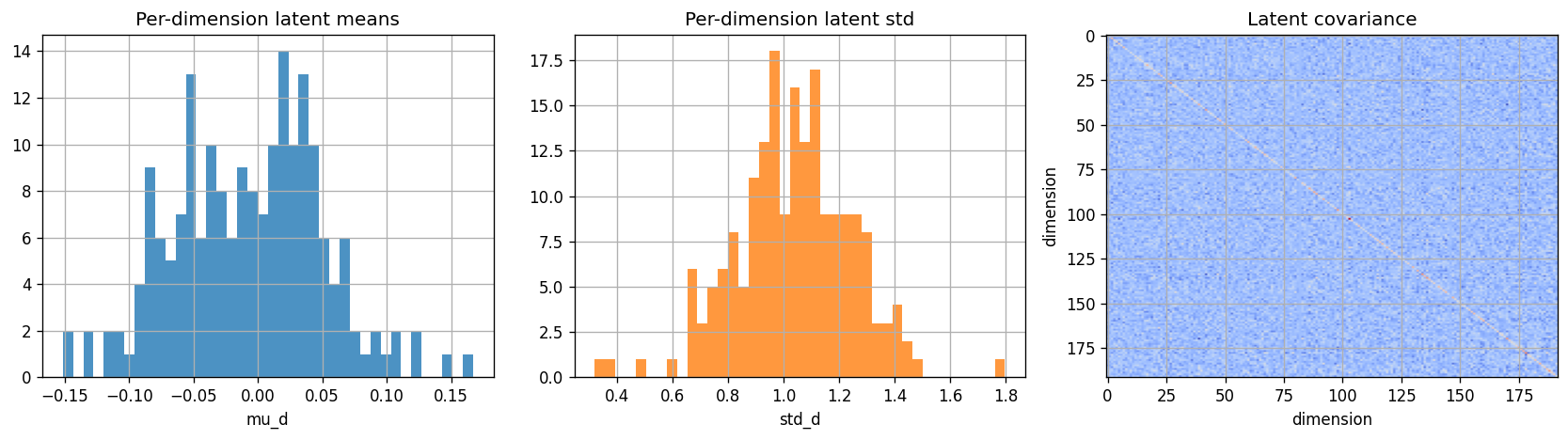}
  \caption{Latent-geometry diagnostics from a fixed sampled subset of \texttt{valsplit}: histogram of per-dimension means, histogram of per-dimension standard deviations, and latent covariance matrix. The figure visually confirms the approximately isotropic geometry summarized numerically in Table~\ref{tab:geometry}.}
  \label{fig:latentdetail}
\end{figure}

Figure~\ref{fig:latentdetail} complements Table~\ref{tab:geometry} with more diagnostics on the same fixed sampled \texttt{valsplit} subset. The per-dimension means remain centered around zero, most per-dimension standard deviations stay near one, and the covariance matrix remains largely diagonal. This is the qualitative signature we expect from the SIGReg objective when the latent space is healthy.

\subsection{Anomaly ranking for temporal QA}
\begin{figure}[!htp]
  \centering
  \includegraphics[width=0.6\linewidth]{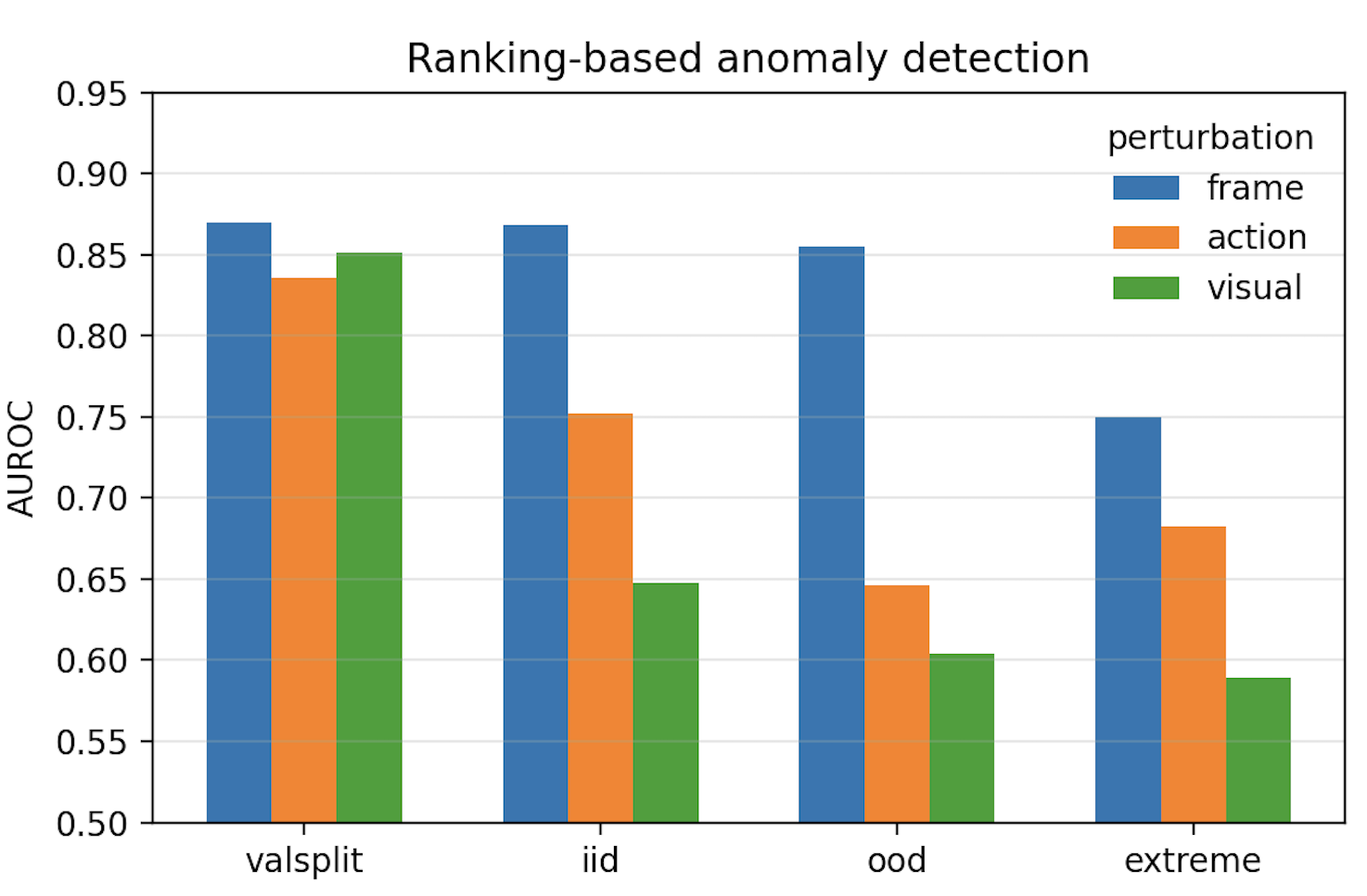}
  \caption{Sampled ranking-based anomaly detection from fixed notebook windows under synthetic future inconsistencies.}
  \label{fig:anomalyobs}
\end{figure}

Figure~\ref{fig:anomalyobs} reports ranking-based anomaly metrics from the fixed sampled notebook diagnostics. Frame-shuffle anomalies are easiest to detect and remain strong on all splits, with AUROC between 0.75 and 0.87. Action-shuffle anomalies are weaker but still meaningfully separable. Visual-shift anomalies are strong on \texttt{valsplit} and clearly weaker on OOD and extreme. For monitoring, this means the latent predictor can serve as a temporal-consistency checker. It is not a semantic anomaly detector in the broad sense; rather, it is a detector for futures that do not fit the recent temporal context under the learned dynamics.

\subsection{Next-step observation quality}
The next operational question is whether the model can estimate if the next acquisition will be useful. We evaluate this both as continuous quality regression and as binary usable-next classification, and we compare LeWM against both persistence and the frozen LightGBM baseline.

For continuous next-step clear/cloud regression, LeWM is best on \texttt{valsplit}, IID, and extreme, with RMSE 0.177, 0.206, and 0.272 versus 0.247, 0.247, and 0.344 for LightGBM and about 0.60 for persistence. OOD is the only split where LightGBM is better on this task (RMSE 0.331 versus 0.361 for LeWM), which reinforces the conclusion that distribution shift remains the main weakness of the latent model.

\begin{figure*}[!htp]
  \centering
  \includegraphics[width=\textwidth]{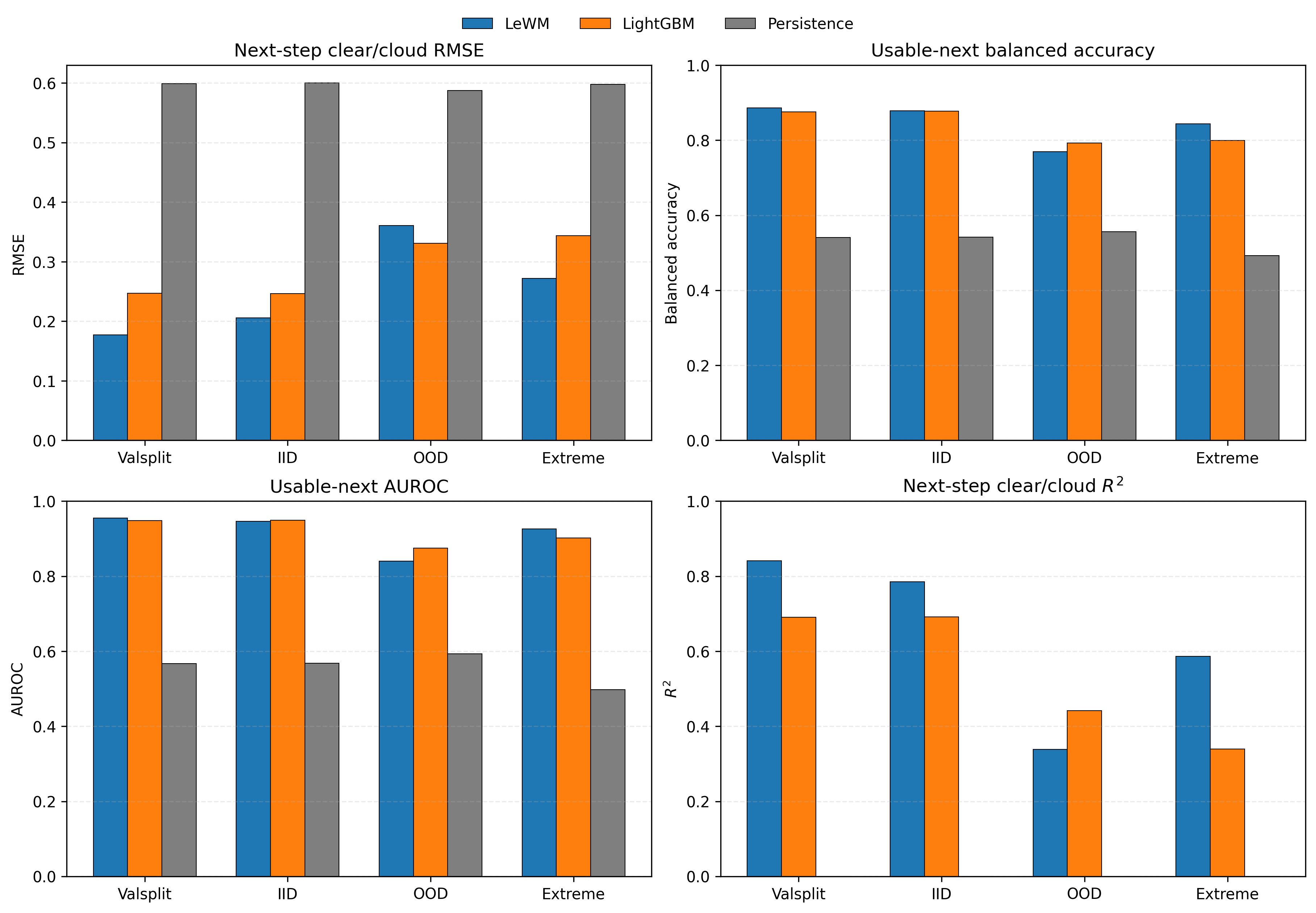}
  \caption{Full frozen-bundle benchmark for next-step observation quality. LeWM is consistently better than persistence and usually better than LightGBM on continuous clear/cloud regression. On binary usable-next classification, LeWM wins on \texttt{valsplit} and extreme, is essentially tied on IID, and loses on OOD.}
  \label{fig:qualitybenchmark}
\end{figure*}

For binary \texttt{usable\_next}, LeWM and LightGBM are close on train-like splits, but the pattern remains task-dependent. LeWM wins on \texttt{valsplit} (balanced accuracy 0.887 versus 0.875) and extreme (0.844 versus 0.799), is effectively tied on IID (0.878 versus 0.878), and loses on OOD (0.769 versus 0.793). Figure~\ref{fig:qualitybenchmark} and Table~\ref{tab:main} summarize this trade-off. The key point is that LeWM clearly beats persistence everywhere, but the stronger tabular baseline remains competitive whenever explicit quality histories already contain much of the signal.

\subsection{Forecasting when usable observations return}
\begin{figure}[!htp]
  \centering
  \includegraphics[width=\linewidth]{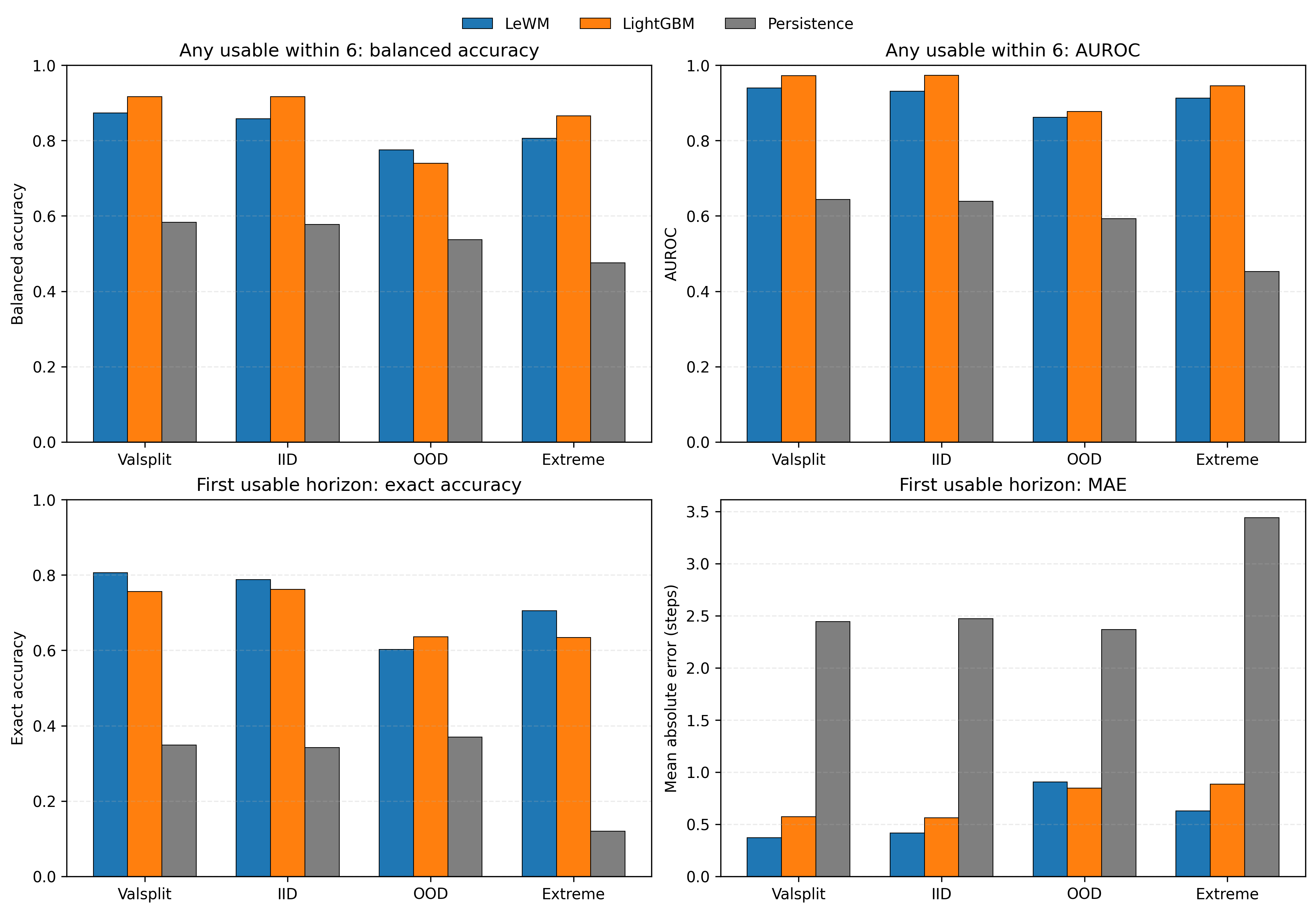}
  \caption{Full frozen-bundle comparison on cloud-aware rollout monitoring tasks. Top row: binary prediction of whether any usable observation occurs within six future steps. Bottom row: exact timing accuracy and mean absolute error for the first usable future observation. LightGBM is strongest on the simpler binary any-usable task on most splits, whereas LeWM is strongest on exact timing on \texttt{valsplit}, IID, and extreme.}
  \label{fig:usable}
\end{figure}

This is the strongest application result of the paper. Given the current context, we ask whether any usable observation will occur within the next six steps and, if so, at what horizon the first usable step will appear (see Table~\ref{tab:main} and Figure~\ref{fig:usable} from the full frozen-bundle benchmark).

For the binary \emph{any-usable-within-six} question, LightGBM is the strongest baseline on \texttt{valsplit}, IID, and extreme. Its balanced accuracy is 0.916, 0.916, and 0.865 on those splits, compared with 0.873, 0.858, and 0.806 for LeWM. OOD is mixed: LeWM has slightly better balanced accuracy (0.775 versus 0.740), while LightGBM has the better AUROC (0.877 versus 0.862). This task therefore behaves like a strong structured-tabular decision problem.

The exact-timing version is where the latent model becomes more convincing. LeWM beats both baselines on \texttt{valsplit}, IID, and extreme, with exact first-usable-horizon accuracies 0.806, 0.787, and 0.705, compared with 0.756, 0.762, and 0.633 for LightGBM and 0.348, 0.341, and 0.120 for persistence. It also substantially reduces mean absolute timing error: 0.372, 0.415, and 0.627 steps for LeWM versus 0.572, 0.563, and 0.885 for LightGBM and 2.45, 2.47, and 3.44 for persistence. OOD is again the one split where LightGBM is better on timing (exact accuracy 0.636 versus 0.602).

\begin{figure}[!htp]
  \centering
  \includegraphics[width=\linewidth]{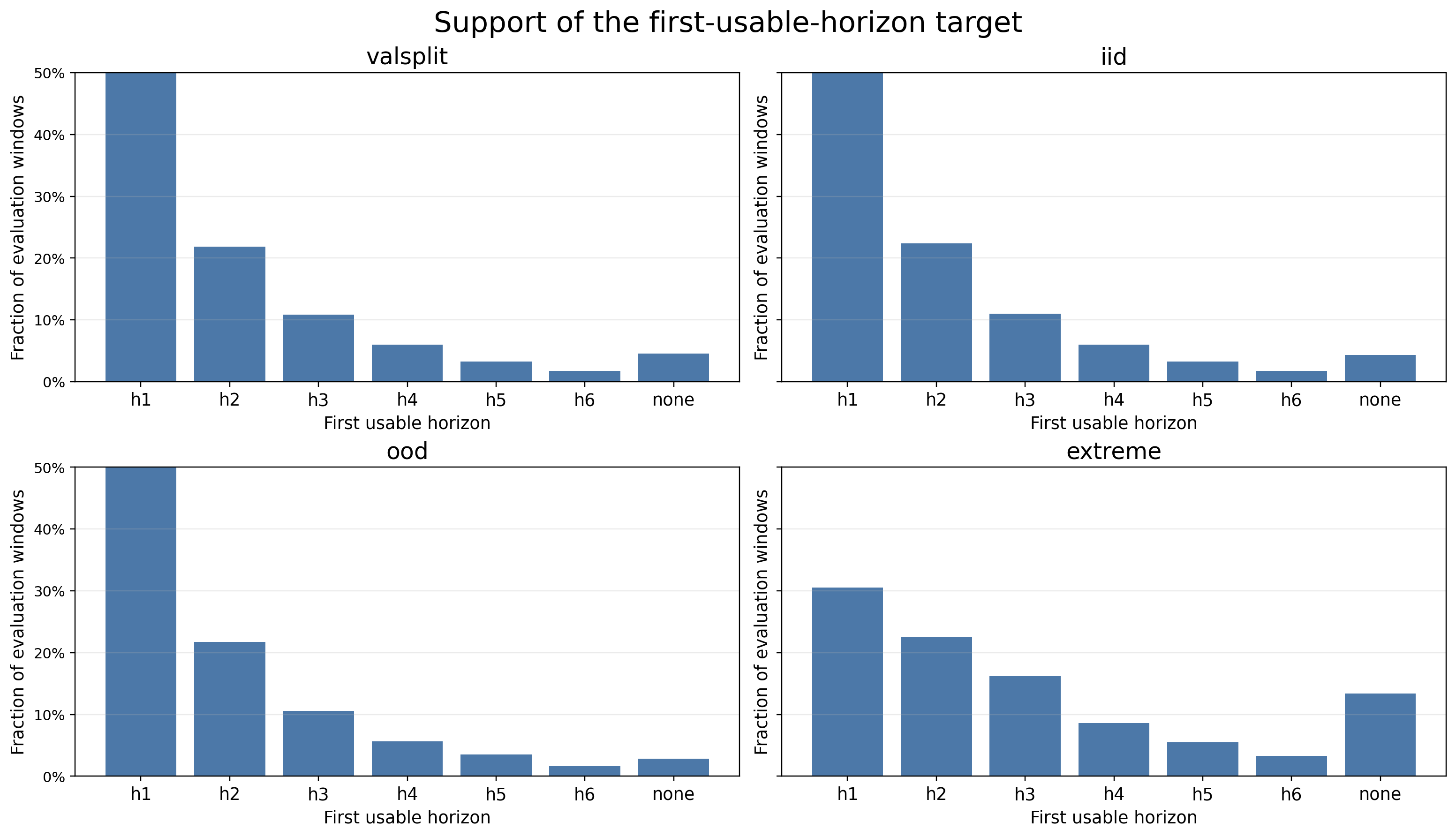}
  \caption{Normalized support of the first-usable-horizon target in the full frozen-bundle evaluation windows. Each panel is normalized within split so the horizon distribution is directly comparable across \texttt{valsplit}, IID, OOD, and extreme. Raw support counts are reported in Table~\ref{tab:appendix-support}. The extreme split is both cloudier and temporally longer, which shifts mass toward later recovery and no recovery within the six-step horizon.}
  \label{fig:support}
\end{figure}

The extreme split is especially informative. It has the lowest clear fraction and the broadest recovery distribution, yet LeWM still recovers a large timing advantage over persistence and a meaningful margin over LightGBM. Figure~\ref{fig:support} from the same full frozen-bundle evaluation shows why this split is harder: its recovery targets are shifted toward later horizons and more no-recovery cases within the six-step window. This is the clearest evidence that the latent model adds value when the target is not just ``will recovery happen?'' but ``when will recovery happen exactly?'' 

\begin{table*}[!htp]
\small
\centering
\scriptsize
\caption{Final observability benchmark on the full frozen-bundle evaluations. Lower is better for RMSE and MAE; higher is better for balanced accuracy, AUROC, and exact timing accuracy. LeWM and LightGBM are each fit once on train windows and frozen for every reported split. Persistence uses the last observed quality state.}
\label{tab:main}
\resizebox{\textwidth}{!}{%
\begin{tabular}{llrrrrrrr}
\toprule
Split & Model & Clear/cloud RMSE & Usable-next bAcc & Usable-next AUROC & Any-usable bAcc & Any-usable AUROC & First-usable exact acc. & First-usable MAE \\
\midrule
Valsplit & LeWM & \textbf{0.177} & \textbf{0.887} & \textbf{0.955} & 0.873 & 0.940 & \textbf{0.806} & \textbf{0.372} \\
 & LightGBM & 0.247 & 0.875 & 0.949 & \textbf{0.916} & \textbf{0.973} & 0.756 & 0.572 \\
 & Persistence & 0.599 & 0.541 & 0.567 & 0.583 & 0.644 & 0.348 & 2.445 \\
\midrule
IID & LeWM & \textbf{0.206} & \textbf{0.878} & 0.947 & 0.858 & 0.931 & \textbf{0.787} & \textbf{0.415} \\
 & LightGBM & 0.247 & \textbf{0.878} & \textbf{0.949} & \textbf{0.916} & \textbf{0.973} & 0.762 & 0.563 \\
 & Persistence & 0.600 & 0.541 & 0.568 & 0.577 & 0.639 & 0.341 & 2.473 \\
\midrule
OOD & LeWM & 0.361 & 0.769 & 0.840 & \textbf{0.775} & 0.862 & 0.602 & 0.907 \\
 & LightGBM & \textbf{0.331} & \textbf{0.793} & \textbf{0.875} & 0.740 & \textbf{0.877} & \textbf{0.636} & \textbf{0.846} \\
 & Persistence & 0.587 & 0.556 & 0.593 & 0.537 & 0.592 & 0.369 & 2.369 \\
\midrule
Extreme & LeWM & \textbf{0.272} & \textbf{0.844} & \textbf{0.927} & 0.806 & 0.912 & \textbf{0.705} & \textbf{0.627} \\
 & LightGBM & 0.344 & 0.799 & 0.902 & \textbf{0.865} & \textbf{0.945} & 0.633 & 0.885 \\
 & Persistence & 0.598 & 0.493 & 0.497 & 0.476 & 0.452 & 0.120 & 3.442 \\
\bottomrule
\end{tabular}%
}
\end{table*}

\subsection{Weather sensitivity}
On fixed sampled notebook windows, perturbing the immediate weather covariates yields coherent but small changes in the predicted next-step usability. Across \texttt{valsplit}, IID, and extreme, drier conditions slightly increase predicted clear fraction and usable score, while wetter conditions decrease them. Temperature perturbations produce smaller effects. OOD sensitivity is near zero. This is a reassuring sanity check: the model responds in the expected direction to exogenous weather forcing, but the effect size is modest and should not be overstated.

\section{Discussion}

\paragraph{Latent forecasting matters, but it is not the main claim.}
The observability results only matter if the model actually learns non-trivial temporal structure. On that point the sampled latent diagnostics are clear: LeWM beats persistence on next-step latent prediction and on six-step rollout across all reported splits. That is the necessary foundation for the cloud-aware observability application but it is not the application itself.

\paragraph{LeWM and LightGBM solve different parts of the problem well.}
The comparison with LightGBM is useful because it shows where the latent model earns its keep. LightGBM is hard to beat on the simpler binary question, ``will any usable observation appear somewhere in the next six steps?'' LeWM, in contrast, is better on continuous clear/cloud regression and on exact first-usable timing on \texttt{valsplit}, IID, and extreme. That suggests the latent representation is most useful when the downstream task depends on a richer temporal state than explicit quality histories and weather covariates can provide on their own.

\paragraph{OOD is still the weak point.}
OOD is the one split where LeWM falls behind more consistently. The geometry diagnostics suggest that this is not just a readout problem. The latent distribution itself shifts away from the train-like regime. For this reason, the cleanest claim is not broad robustness across all deployment conditions; it is strong performance on train-like and difficult-but-related conditions, with a clear drop under stronger distribution shift.

\paragraph{The extreme split is difficult, but it is also informative.}
Extreme is much cloudier and has longer episodes, yet LeWM still holds a clear advantage over persistence and remains stronger than LightGBM on exact recovery timing. That matters because it shows the model is not only exploiting easy near-IID structure. It can still track recovery dynamics in a substantially harsher monitoring regime.

\paragraph{This is a forecasting result, not a planning result.}
\texttt{EarthNet2021} does not provide controllable actions. The covariates are weather drivers. Our perturbation experiments are therefore sensitivity analyses: they show how the model reacts when the forcing changes, but they do not demonstrate policy optimization or counterfactual control.

\paragraph{The current model is global, not spatial.}
LeWM predicts a global future latent derived from the encoder class token. It does not produce future patch tokens or future image maps. That boundary matters. The contribution of this paper is a cloud-aware temporal monitoring model, not a spatial video forecasting system.

\paragraph{Availability of future meteorological covariates.}
All reported experiments condition on the future meteorological covariates provided by \texttt{EarthNet2021}. LeWM does not forecast these covariates; results are conditional observability forecasts under known exogenous drivers.

\section{Conclusion}
We presented a full \texttt{EarthNet2021} application of LeWorldModel built around a change in problem framing: from surface forecasting to \emph{observability forecasting}. The contribution is not a new world-model architecture. It is a careful adaptation, training, and evaluation study that asks a practical EO question: will the next observation be usable, and if not, when is a usable view likely to return?

The experiments support a specific conclusion. In sampled latent diagnostics, LeWorldModel forecasts latent futures far better than persistence on every reported split, which confirms that it learns useful temporal structure. On the full frozen-bundle observability benchmark, those latent forecasts support downstream tasks that matter for cloud-aware monitoring. Compared with persistence, LeWorldModel gives much better next-step usability estimates and much better timing estimates for the return of a usable observation. Compared with a strong frozen LightGBM baseline, it is better on continuous observation-quality regression and on exact recovery timing on \texttt{valsplit}, IID, and extreme, while LightGBM remains stronger on the simpler binary any-usable-within-six task and on OOD more broadly.

Future work should target robustness across geography, season and weather regime. Additionally, the current model gives point predictions which may not be sufficient for operational EO monitoring. Calibrated uncertainty on usable next and first-usable timing would be more useful than a single deterministic forecast.

Appendix~A follows the same order as the results section and collects the full supporting tables and diagnostic figures behind each main-text claim.

\clearpage
\appendix
\section{Supplementary Results}
This appendix follows the same order as the results section in the main paper. Section~\ref{app:train} records the checkpoint used throughout the paper and complements Figure~\ref{fig:training}. Section~\ref{app:latent} expands the latent-forecasting results discussed around Figures~\ref{fig:forecast}, \ref{fig:nextstepdetail}, \ref{fig:extremerollout}, and \ref{fig:latentdetail}, together with Table~\ref{tab:geometry}. Section~\ref{app:anomaly} supports Figure~\ref{fig:anomalyobs}. Section~\ref{app:quality} expands the next-step observation-quality benchmark summarized in Figure~\ref{fig:qualitybenchmark} and Table~\ref{tab:main}. Section~\ref{app:return} expands the recovery tasks summarized in Figures~\ref{fig:usable} and \ref{fig:support} and in Table~\ref{tab:main}. Section~\ref{app:weather} gives the full weather-sensitivity results behind the short main-text discussion.

We keep the same split order throughout: \texttt{valsplit}, IID, OOD, and extreme.

\subsection{Training checkpoint used in all experiments}\label{app:train}
All downstream evaluations in the paper use the same 24-epoch LeWM checkpoint. Table~\ref{tab:appendix-train-final} records the final training and validation metrics for that run and complements the learning-curve summary in Figure~\ref{fig:training}.

\begin{table*}[!htp]
\centering
\small
\caption{Final epoch training metrics of the EarthNet2021 LeWM checkpoint used for all downstream evaluations, transcribed from the final training log.}
\label{tab:appendix-train-final}
\begin{tabular}{lr}
\toprule
Metric & Value \\
\midrule
fit/loss & 0.0814 \\
fit/pred\_loss & 0.0081 \\
fit/sigreg\_loss & 1.4609 \\
validate/loss\_epoch & 0.1079 \\
validate/pred\_loss\_epoch & 0.0164 \\
validate/sigreg\_loss\_epoch & 1.8306 \\
\bottomrule
\end{tabular}
\end{table*}

\subsection{Latent forecasting details}\label{app:latent}
This subsection expands the latent-dynamics results that motivate the use of LeWM as a forecasting model rather than only as a feature extractor. It collects split-wise next-step metrics, rollout metrics broken down by forecast horizon, and latent-geometry figures. All results in this subsection come from fixed sampled diagnostic windows in the executed split notebooks, not from the frozen full-bundle benchmark.

\subsubsection{Next-step latent forecasting}
\begin{figure*}[!htp]
\centering
\includegraphics[width=\textwidth]{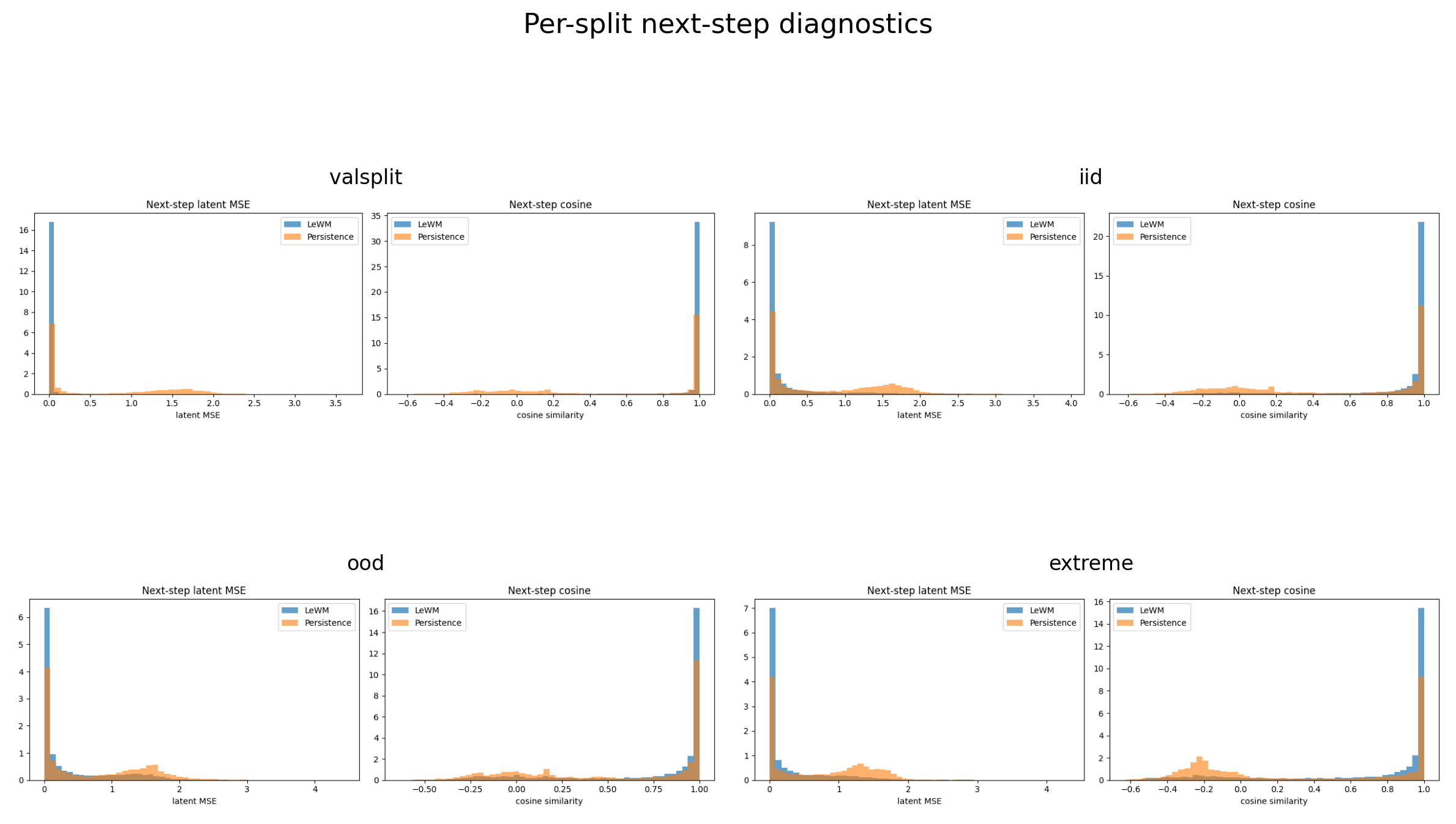}
\caption{Per-split next-step diagnostics from the executed notebooks on fixed sampled windows. For each split we show the full latent-MSE and latent-cosine histograms for LeWM and persistence.}
\label{fig:appendix-nextstep-grid}
\end{figure*}

Table~\ref{tab:appendix-nextstep} gives the split-wise next-step metrics behind the main-text latent-forecasting discussion. Figure~\ref{fig:appendix-nextstep-grid} shows the corresponding distributions for LeWM and persistence.

\begin{table*}[!htp]
\centering
\scriptsize
\caption{Complete next-step latent-forecasting metrics from fixed sampled notebook windows. MSE ratio is model MSE divided by persistence MSE, so lower is better.}
\label{tab:appendix-nextstep}
\resizebox{\textwidth}{!}{%
\begin{tabular}{lrrrrrrrrrrrr}
\toprule
Split & $n$ & pred MSE mean & pred MSE med. & pred MSE p95 & pred cos mean & pred cos med. & pred cos p05 & pers. MSE mean & pers. cos mean & MSE ratio & MSE win & cos win \\
\midrule
Valsplit & 10000 & \textbf{0.015} & 0.003 & 0.030 & \textbf{0.969} & 0.997 & 0.829 & 0.641 & 0.567 & \textbf{0.024} & \textbf{0.829} & \textbf{0.833} \\
IID & 10000 & \textbf{0.188} & 0.020 & 1.170 & \textbf{0.868} & 0.991 & 0.028 & 0.758 & 0.504 & \textbf{0.248} & \textbf{0.743} & \textbf{0.745} \\
OOD & 10000 & \textbf{0.380} & 0.061 & 1.596 & \textbf{0.734} & 0.972 & -0.191 & 0.726 & 0.535 & \textbf{0.524} & \textbf{0.657} & \textbf{0.681} \\
Extreme & 10000 & \textbf{0.283} & 0.028 & 1.357 & \textbf{0.700} & 0.967 & -0.320 & 0.735 & 0.335 & \textbf{0.384} & \textbf{0.646} & \textbf{0.650} \\
\bottomrule
\end{tabular}%
}
\end{table*}

\subsubsection{Six-step rollout by forecast horizon}
\begin{figure*}[!htp]
\centering
\includegraphics[width=\textwidth]{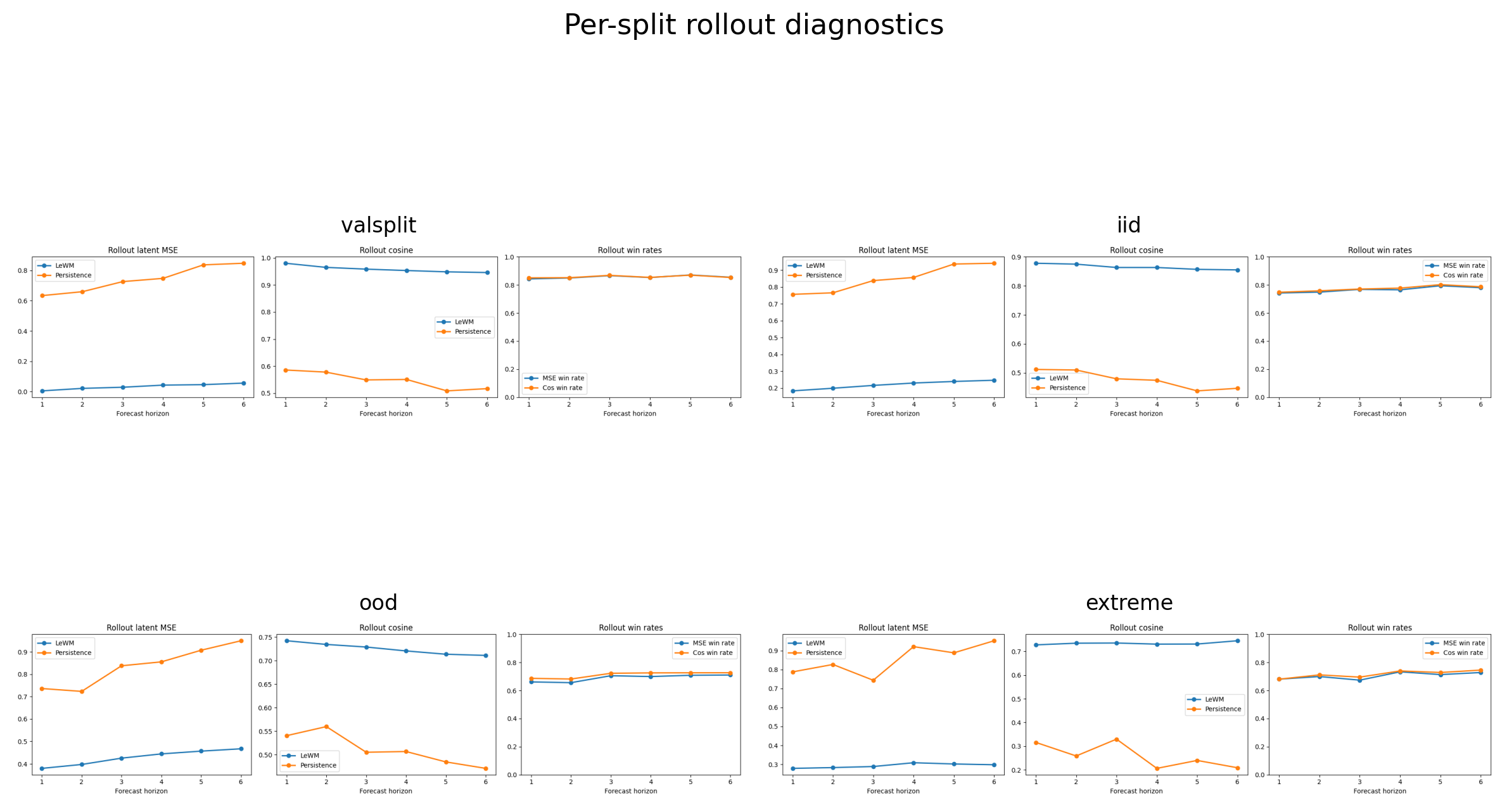}
\caption{Per-split rollout diagnostics from fixed sampled notebook windows. Each split shows horizon-wise latent MSE, cosine similarity, and per-window win rates against persistence.}
\label{fig:appendix-rollout-grid}
\end{figure*}

Tables~\ref{tab:appendix-rollout-valsplit} to \ref{tab:appendix-rollout-extreme} unpack the six-step rollout into per-horizon metrics on the same fixed sampled notebook windows. Figure~\ref{fig:appendix-rollout-grid} summarizes the same comparison visually across all splits.
\begin{table*}[!htp]
\centering
\small
\caption{Six-step rollout metrics on \texttt{valsplit} from fixed sampled notebook windows. Each row corresponds to forecast horizon $h$.}
\label{tab:appendix-rollout-valsplit}
\begin{tabular}{rrrrrrrr}
\toprule
$h$ & pred MSE & pers. MSE & ratio & pred cos & pers. cos & MSE win & cos win \\
\midrule
1 & \textbf{0.004} & 0.634 & \textbf{0.007} & \textbf{0.981} & 0.585 & \textbf{0.843} & \textbf{0.851} \\
2 & \textbf{0.021} & 0.660 & \textbf{0.032} & \textbf{0.965} & 0.578 & \textbf{0.850} & \textbf{0.852} \\
3 & \textbf{0.028} & 0.727 & \textbf{0.039} & \textbf{0.959} & 0.549 & \textbf{0.866} & \textbf{0.868} \\
4 & \textbf{0.042} & 0.748 & \textbf{0.057} & \textbf{0.954} & 0.551 & \textbf{0.853} & \textbf{0.854} \\
5 & \textbf{0.045} & 0.837 & \textbf{0.054} & \textbf{0.949} & 0.508 & \textbf{0.871} & \textbf{0.870} \\
6 & \textbf{0.056} & 0.848 & \textbf{0.066} & \textbf{0.946} & 0.517 & \textbf{0.855} & \textbf{0.853} \\
\bottomrule
\end{tabular}
\end{table*}

\begin{table*}[!htp]
\centering
\small
\caption{Six-step rollout metrics on IID from fixed sampled notebook windows. Each row corresponds to forecast horizon $h$.}
\label{tab:appendix-rollout-iid}
\begin{tabular}{rrrrrrrr}
\toprule
$h$ & pred MSE & pers. MSE & ratio & pred cos & pers. cos & MSE win & cos win \\
\midrule
1 & \textbf{0.183} & 0.756 & \textbf{0.241} & \textbf{0.878} & 0.512 & \textbf{0.743} & \textbf{0.748} \\
2 & \textbf{0.198} & 0.765 & \textbf{0.259} & \textbf{0.875} & 0.510 & \textbf{0.749} & \textbf{0.758} \\
3 & \textbf{0.215} & 0.838 & \textbf{0.257} & \textbf{0.863} & 0.479 & \textbf{0.768} & \textbf{0.771} \\
4 & \textbf{0.229} & 0.856 & \textbf{0.268} & \textbf{0.863} & 0.474 & \textbf{0.765} & \textbf{0.777} \\
5 & \textbf{0.239} & 0.936 & \textbf{0.255} & \textbf{0.857} & 0.438 & \textbf{0.794} & \textbf{0.802} \\
6 & \textbf{0.246} & 0.941 & \textbf{0.261} & \textbf{0.855} & 0.447 & \textbf{0.780} & \textbf{0.787} \\
\bottomrule
\end{tabular}
\end{table*}

\begin{table*}[!htp]
\centering
\small
\caption{Six-step rollout metrics on OOD from fixed sampled notebook windows. Each row corresponds to forecast horizon $h$.}
\label{tab:appendix-rollout-ood}
\begin{tabular}{rrrrrrrr}
\toprule
$h$ & pred MSE & pers. MSE & ratio & pred cos & pers. cos & MSE win & cos win \\
\midrule
1 & \textbf{0.380} & 0.736 & \textbf{0.516} & \textbf{0.742} & 0.540 & \textbf{0.661} & \textbf{0.687} \\
2 & \textbf{0.397} & 0.723 & \textbf{0.549} & \textbf{0.734} & 0.560 & \textbf{0.656} & \textbf{0.682} \\
3 & \textbf{0.426} & 0.837 & \textbf{0.508} & \textbf{0.729} & 0.505 & \textbf{0.706} & \textbf{0.723} \\
4 & \textbf{0.444} & 0.855 & \textbf{0.520} & \textbf{0.721} & 0.507 & \textbf{0.700} & \textbf{0.726} \\
5 & \textbf{0.457} & 0.907 & \textbf{0.504} & \textbf{0.714} & 0.485 & \textbf{0.709} & \textbf{0.726} \\
6 & \textbf{0.467} & 0.949 & \textbf{0.492} & \textbf{0.711} & 0.471 & \textbf{0.710} & \textbf{0.727} \\
\bottomrule
\end{tabular}
\end{table*}

\begin{table*}[!htp]
\centering
\small
\caption{Six-step rollout metrics on extreme from fixed sampled notebook windows. Each row corresponds to forecast horizon $h$.}
\label{tab:appendix-rollout-extreme}
\begin{tabular}{rrrrrrrr}
\toprule
$h$ & pred MSE & pers. MSE & ratio & pred cos & pers. cos & MSE win & cos win \\
\midrule
1 & \textbf{0.279} & 0.788 & \textbf{0.354} & \textbf{0.728} & 0.316 & \textbf{0.681} & \textbf{0.681} \\
2 & \textbf{0.283} & 0.827 & \textbf{0.342} & \textbf{0.735} & 0.259 & \textbf{0.700} & \textbf{0.711} \\
3 & \textbf{0.288} & 0.744 & \textbf{0.388} & \textbf{0.736} & 0.330 & \textbf{0.674} & \textbf{0.696} \\
4 & \textbf{0.308} & 0.921 & \textbf{0.335} & \textbf{0.731} & 0.206 & \textbf{0.733} & \textbf{0.740} \\
5 & \textbf{0.302} & 0.888 & \textbf{0.340} & \textbf{0.731} & 0.240 & \textbf{0.714} & \textbf{0.729} \\
6 & \textbf{0.298} & 0.952 & \textbf{0.313} & \textbf{0.745} & 0.209 & \textbf{0.728} & \textbf{0.746} \\
\bottomrule
\end{tabular}
\end{table*}

\subsubsection{Latent-geometry figures}
\begin{figure*}[!htp]
\centering
\includegraphics[width=\textwidth]{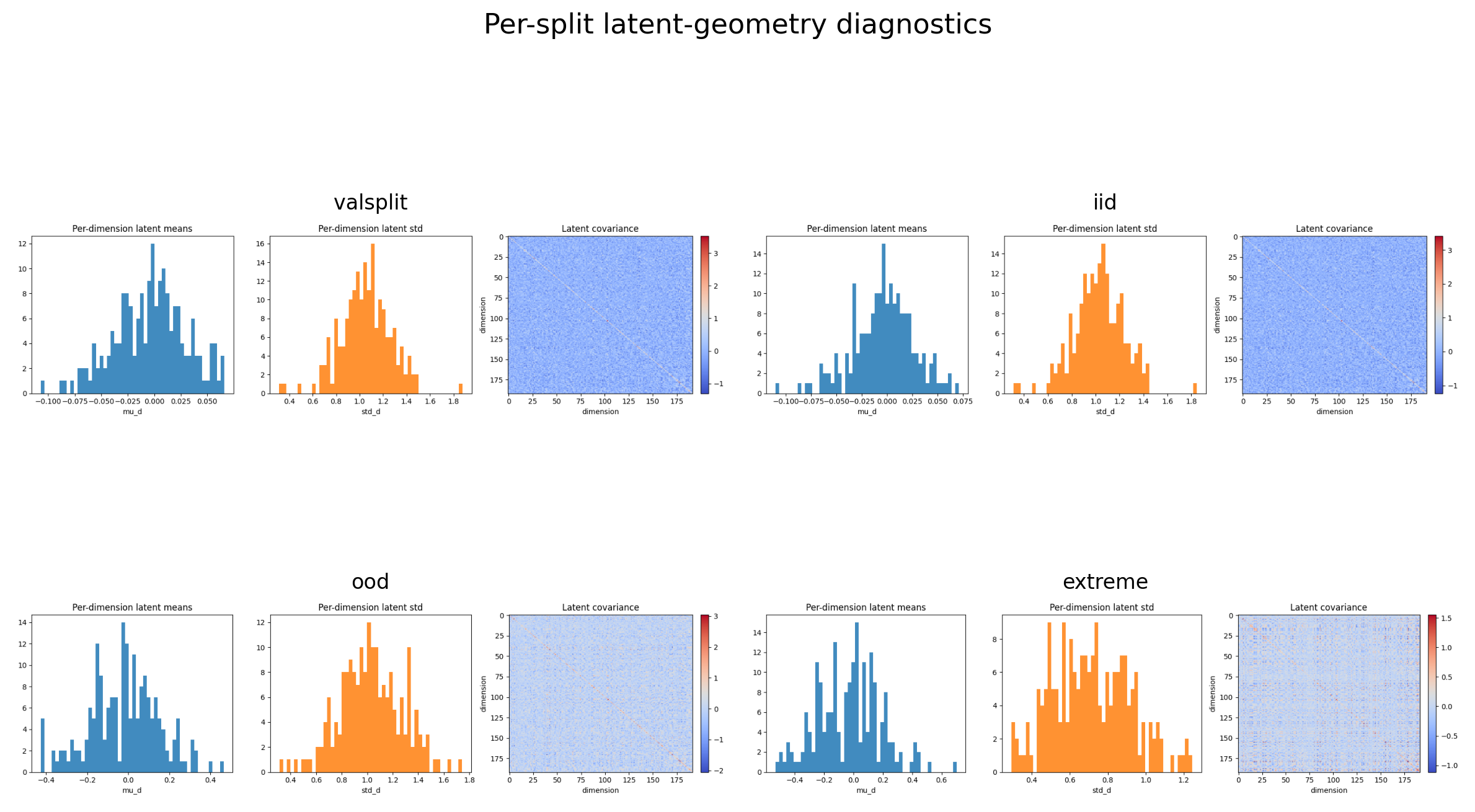}
\caption{Per-split latent-geometry diagnostics from fixed sampled notebook windows. Each panel shows the per-dimension mean histogram, per-dimension standard-deviation histogram, and latent covariance heatmap.}
\label{fig:appendix-latent-grid}
\end{figure*}

Figure~\ref{fig:appendix-latent-grid} complements the scalar geometry summary reported in Table~\ref{tab:geometry}. It shows what the aggregate numbers look like in distribution form, especially the widening gap between train-like splits and OOD or extreme conditions.

\subsection{Temporal-QA anomaly ranking}\label{app:anomaly}
Table~\ref{tab:appendix-anomaly} expands Figure~\ref{fig:anomalyobs}. It reports the anomaly-separation metrics for each synthetic anomaly type and split on the same fixed sampled notebook windows.

\begin{table*}[!htp]
\centering
\small
\caption{Ranking-based anomaly-detection results from the executed notebooks on fixed sampled windows. The score ratio is mean anomaly score divided by mean normal score.}
\label{tab:appendix-anomaly}
\begin{tabular}{llrrr}
\toprule
Split & Anomaly type & Score ratio & AUROC & Avg. precision \\
\midrule
Valsplit & future-frame shuffle & 109.240 & 0.870 & 0.920 \\
Valsplit & future-action shuffle & 23.362 & 0.836 & 0.867 \\
Valsplit & future visual shift & 10.404 & 0.851 & 0.848 \\
IID & future-frame shuffle & 10.974 & 0.868 & 0.910 \\
IID & future-action shuffle & 3.478 & 0.752 & 0.739 \\
IID & future visual shift & 1.681 & 0.648 & 0.608 \\
OOD & future-frame shuffle & 5.788 & 0.855 & 0.891 \\
OOD & future-action shuffle & 1.708 & 0.646 & 0.616 \\
OOD & future visual shift & 1.314 & 0.604 & 0.566 \\
Extreme & future-frame shuffle & 3.888 & 0.750 & 0.788 \\
Extreme & future-action shuffle & 2.119 & 0.683 & 0.655 \\
Extreme & future visual shift & 1.202 & 0.589 & 0.552 \\
\bottomrule
\end{tabular}
\end{table*}

\subsection{Next-step observation quality}\label{app:quality}
This subsection expands the benchmark summarized in Figure~\ref{fig:qualitybenchmark} and Table~\ref{tab:main}. The first table covers continuous clear and cloud regression. The second covers binary \texttt{usable\_next} classification.

\begin{table*}[!htp]
\centering
\small
\caption{Continuous next-step observation-quality regression from the predicted latent. Clear-fraction and cloud-fraction metrics are numerically symmetric in this setup, so one table covers both. All rows come from the full frozen-bundle benchmark.}
\label{tab:appendix-quality-reg}
\begin{tabular}{llrrrrr}
\toprule
Split & Model & RMSE & MAE & Corr. & $R^2$ & RMSE / pers. \\
\midrule
Valsplit & LeWM & \textbf{0.177} & \textbf{0.105} & \textbf{0.917} & \textbf{0.841} & \textbf{0.295} \\
 & LightGBM & 0.247 & 0.187 & 0.840 & 0.691 & 0.412 \\
 & Persistence & 0.599 & 0.435 & 0.085 & -0.822 & 1.000 \\
IID & LeWM & \textbf{0.206} & \textbf{0.124} & \textbf{0.886} & \textbf{0.785} & \textbf{0.343} \\
 & LightGBM & 0.247 & 0.186 & 0.839 & 0.691 & 0.411 \\
 & Persistence & 0.600 & 0.436 & 0.084 & -0.825 & 1.000 \\
OOD & LeWM & 0.361 & \textbf{0.249} & 0.611 & 0.338 & 0.614 \\
 & LightGBM & \textbf{0.331} & 0.264 & \textbf{0.669} & \textbf{0.442} & \textbf{0.564} \\
 & Persistence & 0.587 & 0.420 & 0.118 & -0.757 & 1.000 \\
Extreme & LeWM & \textbf{0.272} & \textbf{0.174} & \textbf{0.778} & \textbf{0.587} & \textbf{0.455} \\
 & LightGBM & 0.344 & 0.296 & 0.660 & 0.339 & 0.575 \\
 & Persistence & 0.598 & 0.439 & 0.002 & -0.996 & 1.000 \\
\bottomrule
\end{tabular}
\end{table*}

\begin{table*}[!htp]
\centering
\small
\caption{Binary next-step usable-observation classification from the full frozen-bundle benchmark. Positive rate refers to the evaluation-set prevalence of usable-next windows.}
\label{tab:appendix-quality-cls}
\begin{tabular}{llrrrrrr}
\toprule
Split & Model & AUROC & Avg. precision & Accuracy@0.5 & Balanced acc.@0.5 & F1@0.5 & Positive rate \\
\midrule
Valsplit & LeWM & \textbf{0.955} & 0.945 & \textbf{0.887} & \textbf{0.887} & \textbf{0.893} & 0.503 \\
 & LightGBM & 0.949 & \textbf{0.949} & 0.875 & 0.875 & 0.876 & 0.503 \\
 & Persistence & 0.567 & 0.563 & 0.542 & 0.541 & 0.582 & 0.503 \\
IID & LeWM & 0.947 & 0.935 & \textbf{0.878} & \textbf{0.878} & \textbf{0.882} & 0.499 \\
 & LightGBM & \textbf{0.949} & \textbf{0.949} & \textbf{0.878} & \textbf{0.878} & 0.878 & 0.499 \\
 & Persistence & 0.568 & 0.558 & 0.541 & 0.541 & 0.579 & 0.499 \\
OOD & LeWM & 0.840 & 0.845 & 0.769 & 0.769 & 0.774 & 0.524 \\
 & LightGBM & \textbf{0.875} & \textbf{0.891} & \textbf{0.793} & \textbf{0.793} & \textbf{0.802} & 0.524 \\
 & Persistence & 0.593 & 0.604 & 0.562 & 0.556 & 0.614 & 0.524 \\
Extreme & LeWM & \textbf{0.927} & 0.803 & \textbf{0.858} & \textbf{0.844} & \textbf{0.759} & 0.275 \\
 & LightGBM & 0.902 & \textbf{0.809} & 0.840 & 0.799 & 0.709 & 0.275 \\
 & Persistence & 0.497 & 0.271 & 0.562 & 0.493 & 0.299 & 0.275 \\
\bottomrule
\end{tabular}
\end{table*}

\subsection{Recovery of usable observations within the rollout horizon}\label{app:return}
This subsection expands the recovery-focused results summarized in Figures~\ref{fig:usable} and \ref{fig:support} and in Table~\ref{tab:main}. We first show the target support by split, then the binary any-usable benchmark, and finally the exact timing benchmark and confusion matrices for the first usable future observation.

\subsubsection{Target support by split}
Table~\ref{tab:appendix-support} gives the raw class counts behind the normalized support plot in Figure~\ref{fig:support} and the normalized confusion matrices in Figure~\ref{fig:appendix-usable-confusions}. It makes the recovery problem concrete by showing how much probability mass sits on immediate recovery, delayed recovery, and no recovery within the six-step horizon.

\begin{table*}[!htp]
\centering
\small
\caption{Support of the first-usable-future target by split in the full frozen-bundle evaluations. ``none'' means that no usable observation appears within the six-step rollout horizon.}
\label{tab:appendix-support}
\begin{tabular}{lrrrrrrr}
\toprule
Split & $h{=}1$ & $h{=}2$ & $h{=}3$ & $h{=}4$ & $h{=}5$ & $h{=}6$ & none \\
\midrule
Valsplit & 18530 & 7831 & 3883 & 2160 & 1179 & 638 & 1634 \\
IID & 32410 & 14147 & 6983 & 3810 & 2044 & 1130 & 2761 \\
OOD & 34192 & 13712 & 6678 & 3570 & 2212 & 1039 & 1807 \\
Extreme & 54959 & 40448 & 29180 & 15484 & 9868 & 5954 & 24107 \\
\bottomrule
\end{tabular}
\end{table*}

\subsubsection{Any-usable-within-six classification}
Table~\ref{tab:appendix-any-usable} gives the full frozen-bundle metrics behind the top row of Figure~\ref{fig:usable} and the corresponding columns of Table~\ref{tab:main}. It shows directly where LeWM, LightGBM, and persistence separate on the simpler binary recovery question.

\begin{table*}[!htp]
\centering
\small
\caption{Binary prediction of whether any usable observation will occur within the six-step rollout horizon. All rows come from the full frozen-bundle benchmark.}
\label{tab:appendix-any-usable}
\begin{tabular}{llrrrrrr}
\toprule
Split & Model & AUROC & Avg. precision & Accuracy & Balanced acc. & F1 & Positive rate \\
\midrule
Valsplit & LeWM & 0.940 & 0.997 & 0.841 & 0.873 & 0.910 & 0.954 \\
 & LightGBM & \textbf{0.973} & \textbf{0.999} & \textbf{0.924} & \textbf{0.916} & \textbf{0.958} & 0.954 \\
 & Persistence & 0.644 & 0.973 & 0.607 & 0.583 & 0.747 & 0.954 \\
IID & LeWM & 0.931 & 0.996 & 0.826 & 0.858 & 0.900 & 0.956 \\
 & LightGBM & \textbf{0.973} & \textbf{0.999} & \textbf{0.924} & \textbf{0.916} & \textbf{0.959} & 0.956 \\
 & Persistence & 0.639 & 0.974 & 0.604 & 0.577 & 0.745 & 0.956 \\
OOD & LeWM & 0.862 & \textbf{0.995} & 0.801 & \textbf{0.775} & 0.887 & 0.971 \\
 & LightGBM & \textbf{0.877} & \textbf{0.995} & \textbf{0.908} & 0.740 & \textbf{0.951} & 0.971 \\
 & Persistence & 0.592 & 0.980 & 0.622 & 0.537 & 0.763 & 0.971 \\
Extreme & LeWM & 0.912 & 0.982 & 0.726 & 0.806 & 0.815 & 0.866 \\
 & LightGBM & \textbf{0.945} & \textbf{0.989} & \textbf{0.916} & \textbf{0.865} & \textbf{0.951} & 0.866 \\
 & Persistence & 0.452 & 0.860 & 0.404 & 0.476 & 0.523 & 0.866 \\
\bottomrule
\end{tabular}
\end{table*}

\subsubsection{Timing of the first usable future observation}
Table~\ref{tab:appendix-timing} gives the exact-timing metrics behind the bottom row of Figure~\ref{fig:usable} and the timing columns of Table~\ref{tab:main}. This is the task where the latent model is most consistently useful on train-like and difficult-but-related splits.

\begin{table*}[!htp]
\centering
\small
\caption{Timing of the first usable future observation on windows where recovery occurs within the six-step horizon. All rows come from the full frozen-bundle benchmark.}
\label{tab:appendix-timing}
\begin{tabular}{llrrrrr}
\toprule
Split & Model & $n$ & Exact acc. & Within-1 acc. & Mean abs. err. & Median abs. err. \\
\midrule
Valsplit & LeWM & 34221 & \textbf{0.806} & \textbf{0.902} & \textbf{0.372} & \textbf{0.000} \\
 & LightGBM & 34221 & 0.756 & 0.855 & 0.572 & \textbf{0.000} \\
 & Persistence & 34221 & 0.348 & 0.493 & 2.445 & 2.000 \\
IID & LeWM & 60524 & \textbf{0.787} & \textbf{0.892} & \textbf{0.415} & \textbf{0.000} \\
 & LightGBM & 60524 & 0.762 & 0.857 & 0.563 & \textbf{0.000} \\
 & Persistence & 60524 & 0.341 & 0.487 & 2.473 & 2.000 \\
OOD & LeWM & 61403 & 0.602 & 0.784 & 0.907 & \textbf{0.000} \\
 & LightGBM & 61403 & \textbf{0.636} & \textbf{0.795} & \textbf{0.846} & \textbf{0.000} \\
 & Persistence & 61403 & 0.369 & 0.511 & 2.369 & 1.000 \\
Extreme & LeWM & 155893 & \textbf{0.705} & \textbf{0.828} & \textbf{0.627} & \textbf{0.000} \\
 & LightGBM & 155893 & 0.633 & 0.764 & 0.885 & \textbf{0.000} \\
 & Persistence & 155893 & 0.120 & 0.239 & 3.442 & 4.000 \\
\bottomrule
\end{tabular}
\end{table*}

\begin{figure*}[!htp]
\centering
\includegraphics[width=0.82\textwidth,height=0.78\textheight,keepaspectratio]{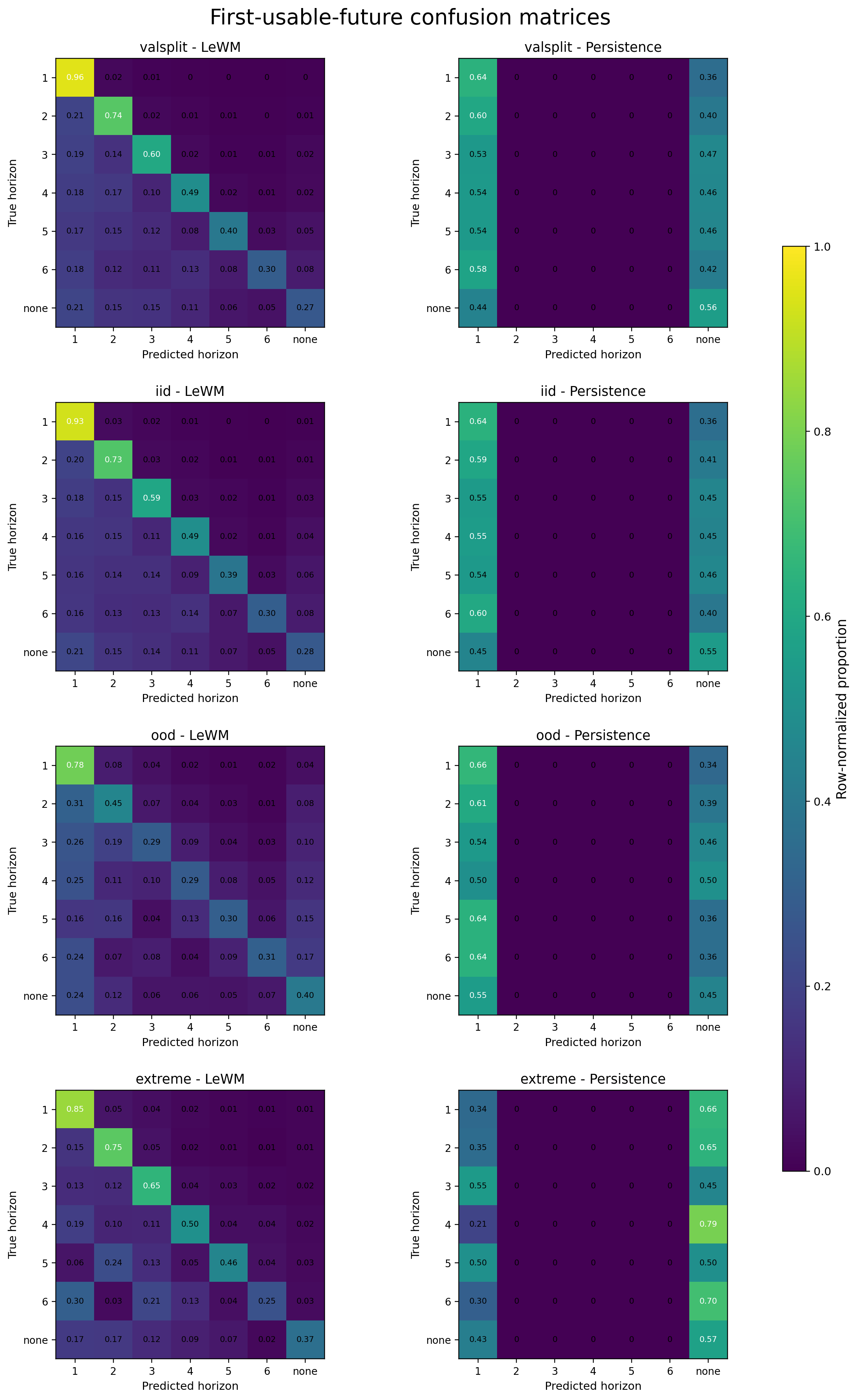}
\caption{Row-normalized confusion matrices for first-usable-future timing from the full frozen-bundle benchmark. Rows are true horizons and columns are predicted horizons, so each row reads as a conditional distribution over predicted recovery times given the true horizon. Raw support counts for each true class are reported in Table~\ref{tab:appendix-support}. The persistence baseline collapses toward predicting horizon 1 or no recovery, whereas LeWM retains informative structure across the full horizon range.}
\label{fig:appendix-usable-confusions}
\end{figure*}

Figure~\ref{fig:appendix-usable-confusions} complements Table~\ref{tab:appendix-timing} by showing where the timing errors occur after row normalization. Table~\ref{tab:appendix-support} should be read alongside it, because the normalized display improves comparability across splits and horizons while the table preserves the raw support counts. The main pattern is that persistence collapses toward very early recovery or no recovery, while LeWM preserves much more of the true horizon structure.

\subsection{Weather sensitivity details}\label{app:weather}
\begin{figure*}[!htp]
\centering
\includegraphics[width=\textwidth]{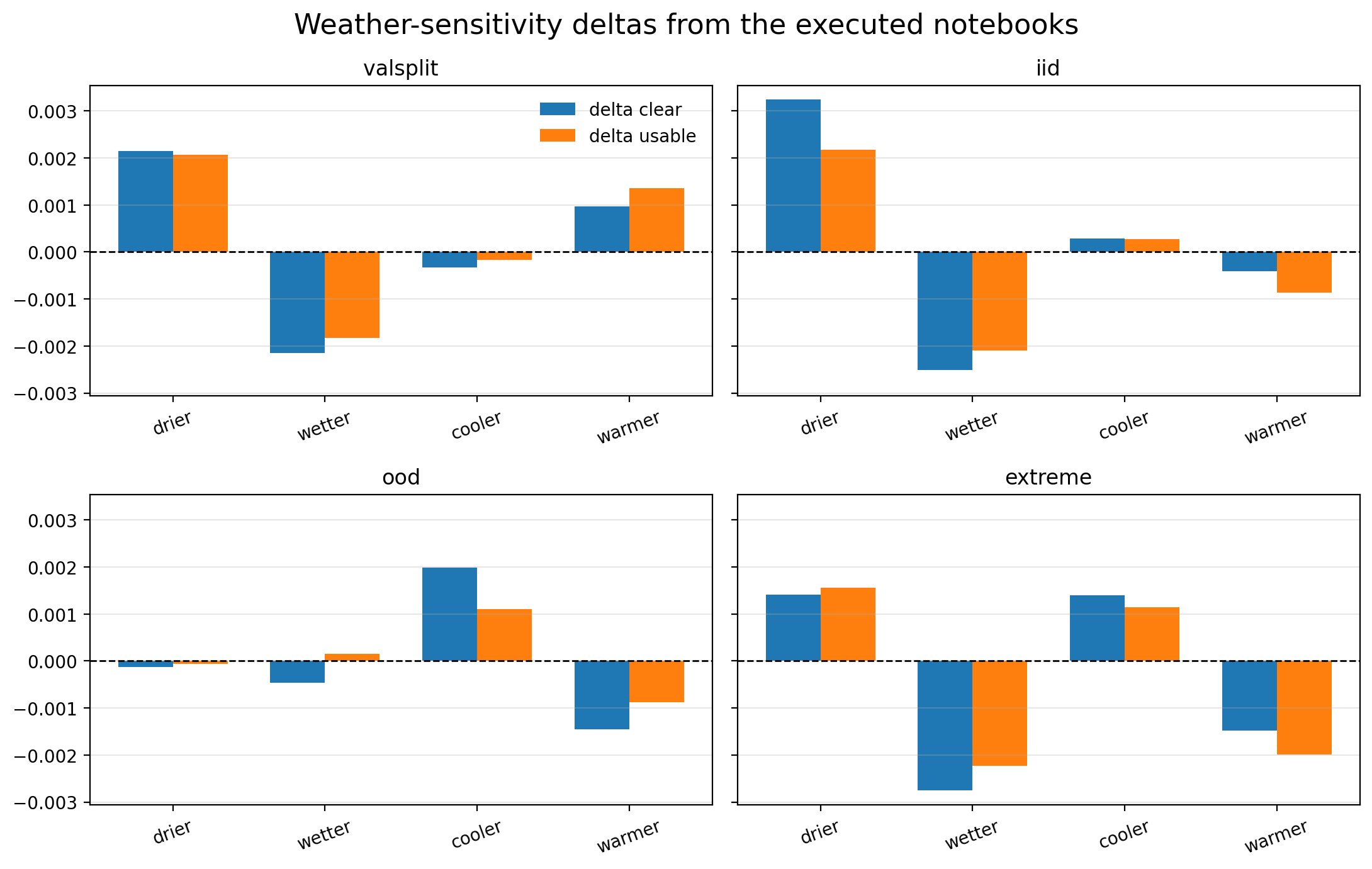}
\caption{Weather-sensitivity deltas from the executed notebooks on fixed sampled windows. Bars show the change in predicted clear fraction and predicted usability relative to the baseline weather input. Effects are coherent but small.}
\label{fig:appendix-weather-grid}
\end{figure*}

This subsection expands the short weather-sensitivity discussion in the main text. All results here come from the fixed sampled notebook windows rather than the frozen full-bundle benchmark. The table gives the split-wise means and deltas, while Figure~\ref{fig:appendix-weather-grid} shows the same effects graphically.

\begin{table*}[!htp]
\centering
\scriptsize
\caption{Weather-sensitivity analysis from fixed sampled notebook windows. We perturb only the immediate future weather covariates and report the mean predicted quality scores and their deltas relative to the unperturbed baseline.}
\label{tab:appendix-weather}
\resizebox{\textwidth}{!}{%
\begin{tabular}{llrrrrrrr}
\toprule
Split & Scenario & Mean pred. clear & Mean pred. cloud & Mean usable score & $\Delta$ clear & $\Delta$ cloud & $\Delta$ usable & frac. more usable \\
\midrule
Valsplit & baseline & 0.494 & 0.506 & 0.411 & 0.0000 & 0.0000 & 0.0000 & -- \\
Valsplit & drier ($-5$ mm rain) & 0.496 & 0.504 & 0.413 & 0.0021 & -0.0021 & 0.0021 & 0.090 \\
Valsplit & wetter ($+5$ mm rain) & 0.492 & 0.508 & 0.409 & -0.0022 & 0.0022 & -0.0018 & 0.074 \\
Valsplit & cooler ($-2^{\circ}$C) & 0.494 & 0.506 & 0.411 & -0.0003 & 0.0003 & -0.0002 & 0.266 \\
Valsplit & warmer ($+2^{\circ}$C) & 0.495 & 0.505 & 0.412 & 0.0010 & -0.0010 & 0.0014 & 0.300 \\
IID & baseline & 0.498 & 0.502 & 0.405 & 0.0000 & 0.0000 & 0.0000 & -- \\
IID & drier ($-5$ mm rain) & 0.501 & 0.499 & 0.407 & 0.0032 & -0.0032 & 0.0022 & 0.089 \\
IID & wetter ($+5$ mm rain) & 0.495 & 0.505 & 0.403 & -0.0025 & 0.0025 & -0.0021 & 0.066 \\
IID & cooler ($-2^{\circ}$C) & 0.498 & 0.502 & 0.405 & 0.0003 & -0.0003 & 0.0003 & 0.295 \\
IID & warmer ($+2^{\circ}$C) & 0.498 & 0.502 & 0.404 & -0.0004 & 0.0004 & -0.0009 & 0.289 \\
OOD & baseline & 0.518 & 0.482 & 0.404 & 0.0000 & 0.0000 & 0.0000 & -- \\
OOD & drier ($-5$ mm rain) & 0.518 & 0.482 & 0.403 & -0.0001 & 0.0001 & -0.0001 & 0.066 \\
OOD & wetter ($+5$ mm rain) & 0.518 & 0.482 & 0.404 & -0.0005 & 0.0005 & 0.0001 & 0.074 \\
OOD & cooler ($-2^{\circ}$C) & 0.520 & 0.480 & 0.405 & 0.0020 & -0.0020 & 0.0011 & 0.356 \\
OOD & warmer ($+2^{\circ}$C) & 0.517 & 0.483 & 0.403 & -0.0015 & 0.0015 & -0.0009 & 0.369 \\
Extreme & baseline & 0.402 & 0.598 & 0.287 & 0.0000 & 0.0000 & 0.0000 & -- \\
Extreme & drier ($-5$ mm rain) & 0.403 & 0.597 & 0.289 & 0.0014 & -0.0014 & 0.0016 & 0.093 \\
Extreme & wetter ($+5$ mm rain) & 0.399 & 0.601 & 0.285 & -0.0028 & 0.0028 & -0.0022 & 0.100 \\
Extreme & cooler ($-2^{\circ}$C) & 0.403 & 0.597 & 0.289 & 0.0014 & -0.0014 & 0.0011 & 0.257 \\
Extreme & warmer ($+2^{\circ}$C) & 0.400 & 0.600 & 0.285 & -0.0015 & 0.0015 & -0.0020 & 0.235 \\
\bottomrule
\end{tabular}%
}
\end{table*}

\newpage
\bibliographystyle{unsrt}
\bibliography{references}

\end{document}